\documentclass{article} 
\usepackage{colm2024_conference}
\usepackage{graphicx}
\usepackage{microtype}
\usepackage{hyperref}
\usepackage{url}
\usepackage{booktabs}
\definecolor{darkblue}{rgb}{0, 0, 0.5}
\hypersetup{colorlinks=true, citecolor=darkblue, linkcolor=darkblue, urlcolor=darkblue}
\usepackage{xspace}
\usepackage{amssymb}
\usepackage{amsmath}
\usepackage{algorithm}
\usepackage{algorithmic}
\usepackage{wrapfig}
\usepackage{tablefootnote}
\usepackage{amssymb}
\usepackage{pifont}
\usepackage{adjustbox}
\usepackage{todonotes}
\usepackage{csquotes}
\usepackage{tikzpagenodes}

\newcommand{\lm}{Granite Code models\xspace}

\usepackage{caption}
\usepackage{subcaption}
\usepackage{algorithm}
\usepackage{algorithmic}
\usepackage{multirow}
\usepackage{xcolor}
\usepackage{fontawesome}

\makeatletter
\newcommand{\makecmd}[2]{%
  \expandafter\newcommand\csname #1\endcsname[1]{%
    #2%
    \def\tempa{##1}%
    \ifx\tempa\@empty
    \else
      -##1B%
    \fi
  }%
}
\makeatother

\makecmd{starcodertwo}{StarCoder2}
\makecmd{starcoderbase}{StarCoderBase}
\makecmd{deepseekcoder}{DeepSeekCoder}
\makecmd{stablecode}{StableCode}
\makecmd{codellama}{CodeLlama}
\makecmd{octocoder}{OctoCoder}
\makecmd{gemma}{Gemma}
\makecmd{codegemma}{CodeGemma}
\makecmd{llamathree}{Llama-3}
\makecmd{mistral}{Mistral}
\makecmd{wizardcoder}{WizardCoder}
\makecmd{llemma}{Llemma}

\makecmd{codellamainstruct}{CodeLlama-Instruct}
\makecmd{deepseekcoderinstruct}{DeepSeekCoder-Instruct}

\title{\centering \LARGE Granite Code Models: A Family of Open \\ Foundation Models for Code Intelligence}

\colmfinalcopy 
\begin{document}
\maketitle

\vspace{0.5cm}
\vspace{-2cm}
\begin{center}

\textbf{Mayank~Mishra}$^\star$\quad
\textbf{Matt~Stallone}$^\star$\quad
\textbf{Gaoyuan~Zhang}$^\star$\quad
\textbf{Yikang~Shen}\quad
\textbf{Aditya~Prasad}\quad
\textbf{Adriana~Meza~Soria}\quad
\textbf{Michele~Merler}\quad
\textbf{Parameswaran~Selvam}\quad
\textbf{Saptha~Surendran}\quad
\textbf{Shivdeep~Singh}\quad
\textbf{Manish~Sethi}\quad
\textbf{Xuan-Hong~Dang}\quad
\textbf{Pengyuan~Li}\quad
\textbf{Kun-Lung~Wu}\quad
\textbf{Syed~Zawad}\quad
\textbf{Andrew~Coleman}\quad
\textbf{Matthew~White}\quad
\textbf{Mark~Lewis}\quad
\textbf{Raju~Pavuluri}\quad
\textbf{Yan~Koyfman}\quad
\textbf{Boris~Lublinsky}\quad
\textbf{Maximilien~de~Bayser}\quad
\textbf{Ibrahim~Abdelaziz}\quad
\textbf{Kinjal~Basu}\quad
\textbf{Mayank~Agarwal}\quad
\textbf{Yi~Zhou}\quad
\textbf{Chris~Johnson}\quad
\textbf{Aanchal Goyal}\quad
\textbf{Hima~Patel}\quad
\textbf{Yousaf~Shah}\quad
\textbf{Petros~Zerfos}\quad
\textbf{Heiko~Ludwig}\quad
\textbf{Asim~Munawar}\quad
\textbf{Maxwell~Crouse}\quad
\textbf{Pavan~Kapanipathi}\quad
\textbf{Shweta~Salaria}\quad
\textbf{Bob~Calio}\quad
\textbf{Sophia~Wen}\quad
\textbf{Seetharami~Seelam}\quad
\textbf{Brian~Belgodere}\quad
\textbf{Carlos~Fonseca}\quad
\textbf{Amith~Singhee}\quad
\textbf{Nirmit~Desai}\quad
\textbf{David~D.~Cox}\quad  \\
\textbf{Ruchir~Puri}$^\dagger$\quad
\textbf{Rameswar~Panda}$^\dagger$\quad

IBM Research \\
$^\star$Equal Contribution \\
$^\dagger$Corresponding Authors \\
\texttt{ruchir@us.ibm.com, rpanda@ibm.com}

\end{center}

\begin{abstract}
Large Language Models (LLMs) trained on code are revolutionizing the software development process. 
Increasingly, code LLMs are being integrated into software development environments to improve the productivity of human programmers, and LLM-based agents are beginning to show promise for handling complex tasks autonomously. Realizing the full potential of code LLMs requires a wide range of capabilities, including code generation, fixing bugs, explaining and documenting code, maintaining repositories, and more. In this work, we introduce the Granite series of decoder-only code models for code generative tasks, trained with code written in 116 programming languages. The \lm family consists of models ranging in size from 3 to 34 billion parameters, suitable for applications ranging from complex application modernization tasks to on-device memory-constrained use cases. 
Evaluation on a comprehensive set of tasks demonstrates that \lm consistently reaches state-of-the-art performance among available open-source code LLMs. The Granite Code model family was optimized for enterprise software development workflows and performs well across a range of coding tasks (e.g. code generation, fixing and explanation), making it a versatile ``all around'' code model. We release all our \lm under an Apache 2.0 license for both research and commercial use.

\ifcolmfinal
\vspace{.75em}
\centering \faGithubSquare~ \url{https://github.com/ibm-granite/granite-code-models}
\else
\fi

\end{abstract}

\section{Introduction}
\label{sec:introduction}

\begin{figure} [h]
\begin{center}
\includegraphics[width=1\linewidth]{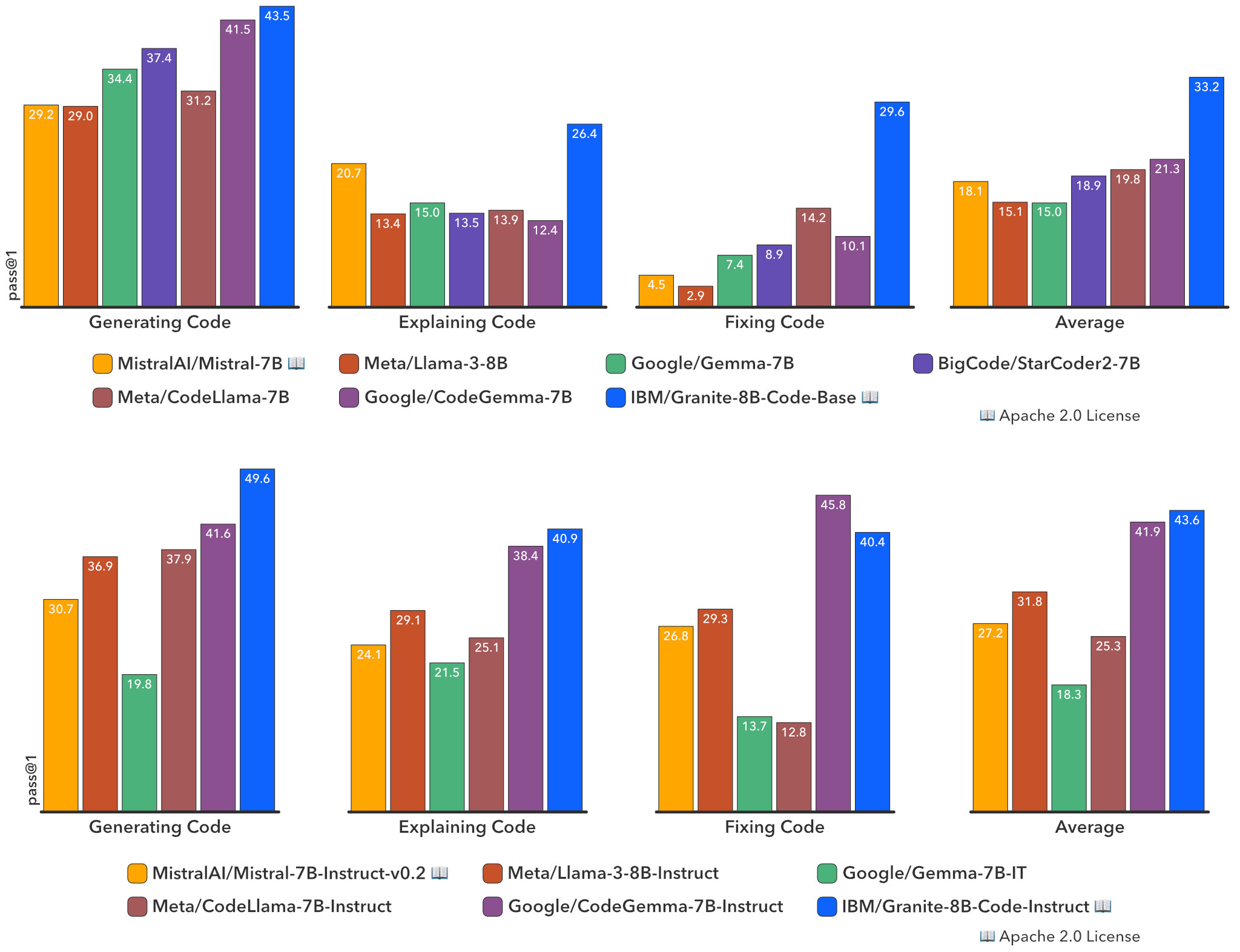}
\end{center} 
\caption{Comparison of Granite-8B-Code (Base/Instruct) with other open source (code) LLMs of similar size on HumanEvalPack~\citep{octopack}, spanning 3 coding tasks and 6 programming languages. See Tables~\ref{tab:hevalsynthesize},\ref{tab:hevalexplain},\ref{tab:hevalfix} for more details.
Best viewed in color.}
\label{fig:compare-8b} \vspace{-2mm}
\end{figure}

Over the last several decades, software has been woven into the fabric of every aspect of our society. As demand for software development surges, it is more critical than ever to increase software development productivity, and LLMs provide promising path for augmenting human programmers. Prominent enterprise use cases for LLMs in software development productivity include code generation, code explanation, code fixing, unit test and documentation generation, application modernization, vulnerability detection, code translation, and more.

Recent years have seen rapid progress in LLM's ability to generate and manipulate code, and a range of models with impressive coding abilities are available today.
Models range in size from single-digit billions of parameters (e.g. Llama-7B~\citep{touvron2023llama}, Gemma-7B~\citep{gemmateam2024gemma}, etc.)  to hundreds of billions: DBRX \citep{dbrx}, Arctic \citep{arctic}, Grok, Mixtral 8x22B~\citep{mixtral8x22b}, Command R+~\citep{commandrplus}, and vary in the generality of intended use, with some models aiming to cover a range of uses outside of code, while others focus primarily on coding-related tasks (e.g. StarCoder~\citep{starcoder,lozhkov2024starcoder}, CodeGen~\citep{codegen}, CodeLlama~\citep{codellama}, and CodeGemma~\citep{codegemma}).

However, there remain important gaps in the current field of LLMs for code, especially in the context of enterprise software development.
First, while very large, generalist LLMs can achieve excellent coding performance, their size makes them expensive to deploy. Smaller code-focused models~\citep{starcoder,lozhkov2024starcoder,codegen,codellama,codegemma} can achieve excellent code generation performance in a smaller and more flexible package, but performance in coding tasks beyond generation (e.g. fixing and explanation) can lag behind code generation performance.

In many enterprise contexts, code LLM adoption can be further complicated by factors beyond the performance of the models. For instance, even open models are sometimes plagued by a lack of transparency about the data sources and data processing methods that went into model, which can erode trust in models in mission critical and regulated contexts. Furthermore, license terms in today's open LLMs can encumber and complicate an enterprise's ability to use a model. 
 
Here, we present \lm, a series of highly capable code LLMs, designed to support enterprise software development across a wide range of coding tasks.
\lm has two main variants that we release in four different sizes (3B, 8B, 20B, and 34B):
\begin{itemize}
    \item \textbf{Granite Code Base:} base foundation models for code-related tasks;
    \item \textbf{Granite Code Instruct:} instruction following models finetuned using a combination of Git commits paired with human instructions and open-source synthetically generated code instruction datasets.
\end{itemize}
The base models in the series have been trained from scratch with a two-phase training strategy.
In phase 1, our model is trained on 3 to 4 trillion tokens sourced from 116 programming languages, ensuring a comprehensive understanding of programming languages and syntax. 
In phase 2, our model is further trained on 500 billion tokens with a carefully designed mixture of high-quality data from code and natural language domains to improve the model's ability to reason. We use the unsupervised language modeling objective to train the base models in both the phases of training. The instruct models are derived by further finetuning the above trained base models on a combination of a filtered variant of CommitPack \citep{octopack}, natural language instruction following datasets (OASST \citep{openassistant}, HelpSteer~\citep{wang2023helpsteer}) and open-source math datasets (MathInstruct~\citep{yue2023mammoth} and MetaMathQA~\citep{yu2023metamath}), including synthetically generated code datasets for improving instruction following and reasoning capabilities.

We conduct extensive evaluations of our code LLMs on a comprehensive set of benchmarks, including HumanEvalPack~\citep{octopack}, MBPP(+)~\citep{mbpp,evalplus}, RepoBench~\citep{liu2023repobench}, ReCode~\citep{recode_wang2022}, and more. 
This set of benchmarks encompasses many different kinds of coding tasks beyond just code synthesis in Python, e.g., code fixing, code explanation, code editing, code translation, etc., across most major programming languages (Python, JavaScript, Java, Go, C++, Rust, etc.).

Our findings reveal that among open-source models, the Granite Code models overall show very strong performance across all model sizes and benchmarks (often outperforming other open-source code models that are twice large compared to Granite). 
As an illustration, figure~\ref{fig:compare-8b} (top) shows a comparison of Granite-8B-Code-Base with other open-source base code LLMs, including recent high-performing general purpose base LLMs like Mistral~\citep{mistral} and LLama-3~\citep{llama3modelcard} on HumanEvalPack~\citep{octopack}. While CodeGemma and StarCoder2 perform reasonably well in generating code, they perform significantly worse on the code fixing and explanation variants of HumanEvalPack. On average, Granite-8B-Code-Base outperforms the most competitive CodeGemma-8B model by almost 12 points on HumanEvalPack (33.2\% vs 21.3\%), despite being trained on significantly less number of tokens (4.5T vs 7.5T tokens).
Besides base models, the instruction tuned variants of our Granite Code models also show strong performance on HumanEvalPack, outperforming other open-source (code) instruction models, demonstrating benefits to a wider set of coding tasks with natural language instructions (see figure~\ref{fig:compare-8b} (bottom)). 

Furthermore, since reasoning is critical for solving complicated questions and tasks, we also test our Granite-8B-Code-Base model on six mathematical benchmarks, including MATH~\citep{cobbe2021training}, GSM8K~\citep{cobbe2021training} and problem solving with access to computational tools, where our Granite 8B model achieves better performance compared to most state-of-the-art 7B or 8B LLMs. 
For example, Granite-8B-Code-Base outperforms Llama-3-8B-Base by $\sim$12 points on GSM8K and $\sim$6 points on MATH (see table~\ref{tab:reasoning}).

The key advantages of \lm include:
\begin{itemize}
    \item \textbf{All-rounder Code LLM}: \lm achieve competitive or state-of-the-art performance on different kinds of code-related tasks, including code generation, explanation, fixing, editing, translation, etc., demonstrating their ability to solve diverse coding tasks;
    \item \textbf{Trustworthy Enterprise-Grade LLM}: All our models are trained on license-permissible data collected following IBM's AI Ethics principles\footnote{\url{https://www.ibm.com/impact/ai-ethics}} and guided by IBM's Corporate Legal team for trustworthy enterprise usage. All the Granite Code models are released under the Apache 2.0 license.
\end{itemize}

We describe our entire data collection, filtering, and preprocessing pipeline in section \ref{sec:data_collection}. Section~\ref{sec:model} describes the details of model architecture, followed by training details in Section~\ref{sec:pretraining}. Section~\ref{sec:instruction_tuning} provides the details about instruction tuning, and Section~\ref{sec:evaluation} describes the experiments and results comparing \lm with other open-source LLMs.
\vspace{-4mm}
\section{Data Collection}
\label{sec:data_collection}
\vspace{-1mm}

In this section, we describe the process of crawling and filtering (Sec.~\ref{sec:crawling}), deduplication (Sec.~\ref{sec:dedup}), HAP/PII filtering (Sec.~\ref{sec:hap}) used to prepare the code data for model training.
We also provide an overview of high-quality natural language data used to enhance the model’s language understanding and mathematical reasoning skills.

\vspace{-1mm}
\subsection{Data Crawling and Filtering}
\label{sec:crawling}
The pretraining code data was sourced from a combination of publicly available datasets like Github Code Clean\footnote{\url{https://huggingface.co/datasets/codeparrot/github-code-clean}}, StarCoderdata\footnote{\url{https://huggingface.co/datasets/bigcode/starcoderdata}}, and additional public code repositories and issues from GitHub. We filter raw data to retain a list of 116 programming languages out of 300+ languages, as listed in Appendix~\ref{sec:lang}. The assignment of data to programming languages is performed based solely on file extension, similar to StarCoder~\citep{starcoder}. 
After language filtering, we
apply four key filtering rules to filter out lower-quality code~\citep{starcoder}: (1) remove files with fewer than 25\% alphabetic characters, (2) except for the XSLT language, filter out files where the string \enquote{\texttt{$<$?xml version=}} appears within the first 100 characters, (3) for HTML files, only keep files where the visible text makes up at least 20\% of the HTML code and has a minimum length of 100 characters, (4) for JSON and YAML files, only keep files that have a character count ranging from 50 to 5000 characters. We also filter GitHub issues using a set of quality metrics that include removing auto-generated text, filtering out non-English issues, excluding comments from bots, and using the number of users engaged in the conversation as an indicator of quality. We also annotate each code file with license information associated with the respective repository, found via Github APIs and only keep files with permissive licenses for model training.

\vspace{-1mm}
\subsection{Exact and Fuzzy Deduplication}
\label{sec:dedup}
We adopt an aggressive deduplication strategy including both exact and fuzzy deduplication to remove documents having (near) identical code content in our training set. For exact deduplication, we first compute SHA256 hash on the document content and remove records having identical hashes.
Post exact deduplication, we apply fuzzy deduplication with the goal of removing code files that may have slight variations and thereby unbiasing the data further. We apply a two-step method for this: (1) compute MinHashes of all the documents and then utilize Locally Sensitive Hashing (LSH) to group
documents based on their MinHash fingerprints, (2) measure Jaccard similarity between each pair of documents in the same bucket and annotate documents except one as duplicates based on a similarity threshold of 0.7. We apply this near-deduplication process to all programming languages including GitHub issues to enhance the richness and diversity of the training dataset.

\vspace{-1mm}
\subsection{HAP, PII, Malware Filtering}
\label{sec:hap}

To reduce the likelihood of generating hateful, abusive, or profane (HAP) language from the models, we make diligent efforts to filter HAP content from the training set. We first create a dictionary of HAP keywords and then annotate each code document with the number of occurrences of such keywords in the content including comments. We filter out documents which exceeds the HAP threshold, computed based on a distributional analysis as well as manual inspection of code files. Moreover, to protect privacy, we follow StarCoder~\citep{starcoder} and make diligent efforts to redact Personally Identifiable Information (PII) from the training set. Specifically, we leverage the StarPII\footnote{\url{https://huggingface.co/bigcode/starpii}} model to detect IP addresses, keys, email addresses, names, user names, and passwords found in the content. The PII redaction step replaces the PII text with the corresponding tokens $\langle$NAME$\rangle$, $\langle$EMAIL$\rangle$, $\langle$KEY$\rangle$, $\langle$PASSWORD$\rangle$ and change the IP address with a synthetically generated IP address, as in \cite{starcoder}.    
We also scan our datasets using ClamAV\footnote{\url{https://www.clamav.net/}} to identify and remove instances of malware in the source code.   

\vspace{-1mm}
\subsection{Natural Language Datasets}
\label{sec:language}

In addition to collecting code data for model training, we curate several publicly available high-quality natural language datasets for improving the model’s proficiency in language understanding and mathematical reasoning. Representative datasets under this category include web documents (Stackexchange, CommonCrawl), mathematical web text (OpenWebMath;~\cite{paster2023openwebmath}, StackMathQA;~\cite{stackmathqa2024}), academic text (Arxiv, Wikipedia), and instruction tuning datasets (FLAN;~\cite{longpre2023flan}, HelpSteer~\citep{wang2023helpsteer}). We do not deduplicate these already preprocessed natural language datasets.
\vspace{-2mm}
\section{Model Architecture}
\label{sec:model_architecture}
\label{sec:model}
\vspace{-1mm}

We train a series of code models of varying sizes based on the transformer decoder architecture \citep{attention}. The model hyperparameters for these models are given in Table \ref{table:architecture-hyperparameters}. For all model architectures, we use pre-normalization \citep{pmlr-v119-xiong20b}: normalization applied to the input of attention and MLP blocks.

\begin{table}[H] \vspace{-1mm}
\caption{Model configurations for Granite Code models.}
    \centering
    \begin{tabular}{lcccc}
    \toprule
    \textbf{Model} & \textbf{3B} & \textbf{8B} & \textbf{20B}  & \textbf{34B} \\
    \midrule
    Batch size & 2048 & 1024 & 576 & 532 \\
    Context length & 2048 & 4096 & 8192 & 8192 \\
    Hidden size & 2560 & 4096 & 6144 & 6144 \\
    FFN hidden size & 10240 & 14336 & 24576 & 24576 \\
    Attention heads & 32 & 32 & 48 & 48 \\
    Key-Value heads & 32 (MHA) & 8 (GQA) & 1 (MQA) & 1 (MQA) \\
    Layers & 32 & 36 &  52 & 88 \\
    Normalization & RMSNorm & RMSNorm & LayerNorm & LayerNorm \\
    Activation & swiglu & swiglu & gelu & gelu \\
    Vocab size & 49152 & 49152 & 49152 & 49152 \\
    \bottomrule
    \end{tabular}
    \label{table:architecture-hyperparameters} \vspace{-2mm}
\end{table}

\noindent\textbf{3B}: The smallest model in the Granite-code model family is trained with  RoPE embedding \citep{su2023roformer} and Multi-Head Attention \citep{attention}. This model use the swish activation function \citep{ramachandran2017searching} with GLU \citep{shazeer2020glu} for the MLP, also commonly referred to as swiglu. For normalization, we use RMSNorm \citep{zhang2019root} since it's computationally more efficient than LayerNorm \citep{ba2016layer}. The 3B model is trained with a context length of 2048 tokens.

\noindent\textbf{8B}: The 8B model has a similar architecture as the 3B model with the exception of using Grouped-Query Attention (GQA) \citep{ainslie2023gqa}. Using GQA offers a better tradeoff between model performance and inference efficiency at this scale. We train the 8B model with a context length of 4096 tokens.

\noindent\textbf{20B}: The 20B code model is trained with learned absolute position embeddings. We use Multi-Query Attention \citep{shazeer2019fast} during training for efficient downstream inference. For the MLP block, we use the GELU activation function \citep{hendrycks2023gaussian}. For normalizing the activations, we use LayerNorm \citep{ba2016layer}. This model is trained with a context length of 8192 tokens.

\begin{figure}
\begin{center}
\includegraphics[width=0.9\linewidth]{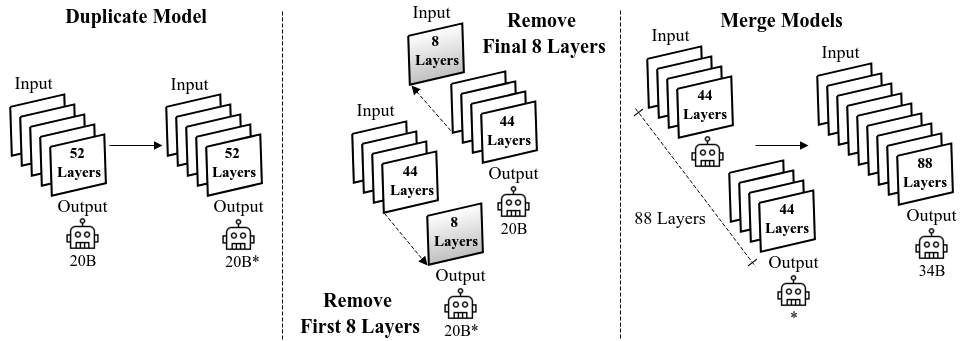}
\end{center} 
\caption{An overview of depth upscaling~\citep{kim2024solar} for efficient training of Granite-34B-Code. We utilize the 20B model after 1.6T tokens to start training of 34B model with the same code pretraining data without any changes to the training and inference framework.}
\label{fig:solar} \vspace{-2mm}
\end{figure}

\noindent\textbf{34B}: To train the 34B model, we follow the approach by \citeauthor{kim2024solar} for depth upscaling of the 20B model. 
Specifically, we first duplicate the 20B code model with 52 layers and then remove final 8 layers from the original model and initial 8 layers from its duplicate to form two models. Finally, we concatenate both models to form Granite-34B-Code model with 88 layers (see Figure~\ref{fig:solar} for an illustration).
After the depth upscaling, we observe that the drop in performance compared to 20B model is pretty small contrary to what is observed by \citeauthor{kim2024solar}. This performance is recovered pretty quickly after we continue pretraining of the upscaled 34B model. Similar, to 20B, we use a 8192 token context during pretraining.

\vspace{-1mm}
\section{Pretraining}
\label{sec:pretraining}

In this section, we provide details on two phase training (Sec.~\ref{sec:phase-training}), training objectives (Sec.~\ref{sec:loss}), optimization (Sec.~\ref{sec:optimization}) and infrastructure (Sec.~\ref{sec:infra}) used in pretraining the models. 

\vspace{-1mm}
\subsection{Two Phase Training}
\label{sec:phase-training}

Granite Code models are trained on 3.5T to 4.5T tokens of code data and natural language datasets related to code. Data is tokenized via byte pair encoding (BPE,~\citep{sennrich2015neural}), employing the same tokenizer as StarCoder~\citep{starcoder}. Following~\citep{shen2024jetmoe,hu2024minicpm}, we utilize high-quality data with two phases of training as follows.

\begin{itemize}
    \item \textbf{Phase 1 (code only training)}: During phase 1, both 3B and 8B models are trained for 4 trillion tokens of code data comprising 116 languages. The 20B parameter model is trained on 3 trillion tokens of code. The 34B model is trained on 1.4T tokens after the depth upscaling which is done on the 1.6T checkpoint of 20B model.
    \item \textbf{Phase 2 (code + language training)}: In phase 2, we include additional high-quality publicly available data from various domains, including technical, mathematics, and web documents, to further improve the model’s performance in reasoning and problem solving skills, which are essential for code generation. We train all our models for 500B tokens (80\% code and 20\% language data) in phase 2 training.  
\end{itemize}

\vspace{-1mm}
\subsection{Training Objective}
\label{sec:loss}
For training of all our models, we use the causal language modeling objective and Fill-In-the-Middle (FIM) \citep{bavarian2022efficient} objective. The FIM objective is tasked to predict inserted tokens with the given context and subsequent text. We train our models to work with both PSM (Prefix-Suffix-Middle) and SPM (Suffix-Prefix-Middle) modes, with relevant formatting control tokens, same as StarCoder~\citep{starcoder}.

The overall loss is computed as a weighted combination of the 2 objectives:
\begin{equation}
    \mathcal{L} = \alpha \mathcal{L}_{CLM} + (1 - \alpha) \mathcal{L}_{FIM}
\end{equation}
We emperically set $\alpha = 0.5$ during training and find that this works well in practice leading to SOTA performance on both code completion and code infilling tasks. It should be noted that the FIM objective is only used during pretraining, however we drop it during instruction finetuning i.e we set $\alpha = 1$.

\vspace{-1mm}
\subsection{Optimization}
\label{sec:optimization}
We use AdamW optimizer \citep{adam} with $\beta_1 = 0.9, \beta_2 = 0.95$ and weight decay of 0.1 for training all our Granite code models. For the phase-1 pretraining, the learning rate follows a cosine schedule starting from $3 \times 10^{-4}$ which decays to $3 \times 10^{-5}$ with an initial linear warmup step of 2k iterations. For phase-2 pretraining, we start from $3 \times 10^{-4}$ ($1.5 \times 10^{-4}$ for 20B and 34B models) and adopt an exponential decay schedule to anneal it to 10\% of the initial learning rate.
We use a batch size of 4M-5M tokens depending on the model size during both phases of pretraining.

To accelerate training, we use FlashAttention 2 \citep{flash, flash2}, the persistent layernorm kernel, Fused RMSNorm kernel (depending on the model) and the Fused Adam kernel available in \href{https://github.com/NVIDIA/apex}{NVIDIA's Apex library}. We use a custom fork of NVIDIA's Megatron-LM \citep{megatron1, megatron2} for distributed training of all our models. We train with a mix of 3D parallelism: tensor parallel, pipeline parallel and data parallel. We also use sequence parallelism \citep{sequence-parallel} for reducing the activation memory consumption of large context length during training. We use Megatron's distributed optimizer with mixed precision training \citep{mixed-precision} in BF16 \citep{bf16} with gradient all-reduce and gradient accumulation in FP32 for training stability.

\vspace{-1mm}
\subsection{Infrastructure}
\label{sec:infra}

We train the Granite Code models using IBM's two supercomputing clusters, namely Vela and Blue Vela, outfitted with NVIDIA A100 and H100 GPUs, respectively. In the Vela A100 GPU cluster, each node has 2$\times$ Intel Xeon Scalable Processors with 8$\times$ 80GB A100 GPUs connected to each other by NVLink and NVSwitch. The Vela cluster adopts RoCE (RDMA over Converged Ethernet) and GDR (GPU-direct RDMA) for high-performance networking. Similarly, each node in Blue Vela cluster consists of dual 48-core Intel processors with 8× 80GB H100 GPUs. Blue Vela employs 3.2Tbps InfiniBand interconnect to facilitate seamless communication between nodes, known for their high throughput and low latency. In addition, Blue Vela employs a separate, dedicated InfiniBand Storage fabric providing 800Gbps per compute node, backed by multiple ESS6000 storage appliances. Both clusters provide a scalable and efficient infrastructure for training our models over thousands of GPUs. We estimate the carbon emissions from pretraining the Granite Code models to be $\sim$455 t$\mathrm{CO_2}$eq, which is computed based on the total energy usage in the models and US national average carbon intensity factor of 0.423 kg $\mathrm{CO_2}$eq/KWh without taking the location of data centers in consideration. The Blue Vela cluster runs on 100\% renewable energy to minimize the environmental impact.

\vspace{-1mm}
\section{Instruction Tuning}
\label{sec:instruction_tuning}

Finetuning code LLMs on a variety of tasks explained via instructions has been shown to improve model usability and general performance. While there has been much progress in code instruction tuning, most of them adopt synthetically generated data from OpenAI models, which limits the model use in many enterprise applications. Thus, following OctoCoder~\citep{octopack}, we use only a combination of permissively licensed data, with an aim to enhance instruction following capabilities of our models, including logical reasoning and problem-solving skills. Specifically, Granite Code Instruct models are trained on the following types of data.

\begin{itemize}
    \item \textbf{Code Commits Dataset}: CommitPackFT~\citep{octopack}, a filtered version of full CommitPack dataset across 92 programming languages\footnote{We selected 92 programming languages that are common across the original CommitPackFT and our list of 116 languages used during pretraining.}; 
    \item \textbf{Math Datasets}: MathInstruct\footnote{We removed GSM8K-RFT and Camel-Math from MathInstruct due to unknown or NC license.}~\citep{yue2023mammoth} and MetaMathQA~\citep{yu2023metamath}; 
    \item \textbf{Code Instruction Datasets}: Glaive-Code-Assistant-v3\footnote{\url{https://huggingface.co/datasets/glaiveai/glaive-code-assistant-v3}}, Self-OSS-Instruct-SC2\footnote{\url{https://huggingface.co/datasets/bigcode/self-oss-instruct-sc2-exec-filter-50k}}, Glaive-Function-Calling-v2\footnote{\url{https://huggingface.co/datasets/glaiveai/glaive-function-calling-v2}}, NL2SQL\footnote{\url{https://huggingface.co/datasets/bugdaryan/sql-create-context-instruction}} and few synthetically generated API calling datasets~\citep{basu2024api}; 
    \item \textbf{Language Instruction Datasets}: High-quality datasets like HelpSteer~\citep{wang2023helpsteer}, an open license-filtered version of Platypus\footnote{\url{https://huggingface.co/datasets/garage-bAInd/Open-Platypus}}~\citep{platypus2023} including a collection of hardcoded prompts to ensure model generates correct outputs given inquiries about its name or developers.
\end{itemize}

For training, we use a cosine scheduler with 250 warmup steps, an initial learning rate $10^{-5}$, and train for three epochs. Further, we add random, uniform noise with a magnitude of $\frac{5}{\sqrt{Nh}}$, where $N$ is the sequence length and $h$ is the embedding dimension, to the embedding vector, as proposed by \citeauthor{neftune}. The additional noise improved overall answer quality of the instruction model. 
We use FlashAttention 2 \citep{flash2, flash} with a Padding-Free Transformer\footnote{\url{https://huggingface.co/blog/mayank-mishra/padding-free-transformer}} implementation to reduce GPU memory usage and redundant FLOPs during finetuning. 
We also use full activation checkpointing \citep{sequence-parallel}, which allows us to finetune our Granite-20B-Code models with 8K context length within a single node within a few hours on 8$\times$A100 GPUs.
\vspace{-1mm}
\section{Evaluation}
\label{sec:evaluation}

\begin{table}[t]
    \caption{Summary of evaluation tasks.} 
    \label{tab:eval_summary}
    \centering
    \begin{tabular}{ccc}
    \toprule
    \textbf{Task}             & \textbf{Benchmark}             & 
\textbf{Reference}        \\
    \midrule
    Multilingual code generation & HumanEvalSynthesize & \cite{octopack}\\
    Multilingual code generation & MultiPL-E & \cite{cassano2023multipl}\\
    Python code generation & MBPP  & \cite{mbpp} \\
    Python code generation & MBPP+  & \cite{evalplus}\\
    Data science code generation & DS1000 & \cite{lai2023ds} \\
    Repository-level code generation & RepoBench & \cite{liu2023repobench}\\
    Repository-level code generation & CrossCodeEval & \cite{ding2023crosscodeeval}\\
    Fill-in-the-middle code completion & SantaCoder-FIM & \cite{allal2023santacoder}\\
    Multilingual code explanation & HumanEvalExplain & \cite{octopack}\\
    Multilingual code fixing & HumanEvalFix & \cite{octopack}\\
    Code editing & CanItEdit & \cite{cassano2024edit} \\
    Code translation & CodeLingua & \cite{pan_lost_2024}\\
    Code execution & CruxEval & \cite{gu2024cruxeval} \\
    Math reasoning & MATH & \cite{hendrycksmath2021}\\
    Math reasoning & GSM8K & \cite{cobbe2021training}\\
    Math reasoning & SAT & \cite{azerbayev2023llemma} \\
    Math reasoning & OCW & \cite{lewkowycz2022solving} \\
    Function calling & BFCL & \cite{berkeley-function-calling-leaderboard} \\
    Model robustness & ReCode & \cite{recode_wang2022}\\
    \bottomrule
    \end{tabular}
\end{table}

We evaluate Granite Code models on a wide variety of tasks, including code generation, code explanation, code fixing, code editing, math reasoning, etc., as shown in Table~\ref{tab:eval_summary}. We compare our models with several open-source code LLMs: StableCode~\citep{pinnaparaju2024stable}, Code Llama~\citep{roziere2023code}, StarCoder~\citep{li2023starcoder}, StarCoder2~\citep{lozhkov2024starcoder}, and CodeGemma\footnote{\url{https://storage.googleapis.com/deepmind-media/gemma/codegemma_report.pdf}}, including recent high-performing general purpose open LLMs like Mistral~\citep{jiang2023mistral} and LLama-3\footnote{\url{https://github.com/meta-llama/llama3}}.
For all the benchmarks, we evaluate the baseline models (including ours) using the same script and environment for fair comparison.

\subsection{Code Generation}

\begin{table}[th]
    \caption{Pass@1 performance on HumanEvalSynthesize benchmark~\citep{octopack}. All models are evaluated using greedy decoding with completion format for the base models, and instruction template for the instruction-tuned models.
    }
    \label{tab:hevalsynthesize}
    \centering
    \small
    \resizebox{\linewidth}{!}{\begin{tabular}{cc|cccccc|c}
        \toprule
        \textbf{Model} & \textbf{Prompt} & \textbf{Python} & \textbf{JavaScript} & \textbf{Java} & \textbf{Go} & \textbf{C++} & \textbf{Rust} & \textbf{Avg.} \\
        \midrule
        \multicolumn{9}{c}{Base Models} \\
        \midrule
        \starcoderbase{3} & Completion & 25.6 & 22.6 & 24.4 & 18.3 & 23.2 & 16.5 & 21.8 \\
        \stablecode{3} & Completion & 24.4 & 32.3 & 34.1 & 21.3 & 33.5 & 21.3 & 27.8 \\
        \starcodertwo{3} & Completion & 27.4 & 36.0 & \textbf{42.1} & 23.8 & \textbf{36.6} & \textbf{24.4} & 31.7 \\
        \codegemma{2} & Completion & \textbf{39.0} & \textbf{37.8} & 37.8 & 13.4 & 33.5 & 20.7 & 30.4 \\
        Granite-3B-Code-Base & Completion & 36.6 & 37.2 & 40.9 & \textbf{26.2} & 35.4 & 22.0 & \textbf{33.1} \\
        \midrule
        \starcoderbase{7} & Completion & 26.8 & 28.7 & 31.7 & 22.6 & 28.0 & 22.6 & 26.7 \\
        \codellama{7} & Completion & 35.4 & 36.0 & 39.0 & 21.3 & 31.1 & 24.4 & 31.2 \\
        \starcodertwo{7} & Completion & 38.4 & 43.3 & 48.2 & \textbf{31.7} & 38.4 & 24.4 & 37.4 \\
        \codegemma{7} & Completion & 41.5 & 48.8 & 54.9 & 26.8 & \textbf{44.5} & 32.3 & 41.5 \\
        Granite-8B-Code-Base & Completion & \textbf{43.9} & \textbf{52.4} & \textbf{56.1} & \textbf{31.7} & 43.9 & \textbf{32.9} & \textbf{43.5} \\
        \midrule
        \starcoderbase{15} & Completion & 32.3 & 36.6 & 40.2 & 25.6 & 31.1 & 25.6 & 31.9 \\
        \codellama{13} & Completion & 41.5 & 42.7 & 51.8 & 26.8 & 40.9 & 23.2 & 37.8 \\
        \starcodertwo{15} & Completion & 44.5 & 47.0 & 51.8 & \textbf{33.5} & \textbf{50.0} & \textbf{39.6} & \textbf{44.4} \\
        Granite-20B-Code-Base & Completion & \textbf{48.2} & \textbf{50.0} & \textbf{59.1} & 32.3 & 40.9 & 35.4 & 44.3 \\
        \midrule
        \codellama{34} & Completion & 47.4 & 48.2 & 45.6 & 34.1 & 47.0 & 37.2 & 43.3 \\
        Granite-34B-Code-Base & Completion & \textbf{48.2} & \textbf{54.9} & \textbf{61.6} & \textbf{40.2} & \textbf{50.0} & \textbf{39.6} & \textbf{49.1} \\
        \codellama{70} & Completion & 55.5 & 55.5 & 65.2 & 40.9 & 55.5 & 43.9 & 52.8 \\
        \midrule
        \gemma{2} & Completion & 20.1 & 23.2 & 19.5 & 13.4 & 18.3 & 8.5 & 17.2 \\
        \gemma{7} & Completion & 33.5 & 41.5 & 43.9 & 26.2 & 35.4 & 26.2 & 34.4 \\
        \mistral{7}-v0.2 & Completion & 32.9 & 34.1 & 36.6 & 22.6 & 30.5 & 18.3 & 29.2 \\
        Mixtral-8x7B-v0.1 & Completion & 42.1 & 53.7 & 52.4 & 33.5 & 42.7 & 35.4 & 43.3 \\
        Mixtral-8x22B-v0.1 & Completion & 51.2 & 57.9 & 64.6 & 40.9 & 57.3 & 32.9 & 50.8 \\
        \llamathree{8} & Completion & 26.2 & 37.8 & 40.2 & 11.0 & 37.2 & 21.3 & 29.0 \\
        \llamathree{70} & Completion & \hspace{1.5mm}25.0$^\star$ & 51.2 & 62.2 & 21.3 & 53.7 & 37.8 & 41.9 \\
        \midrule
        \multicolumn{9}{c}{Instruct Models} \\
        \midrule
        \codegemma{7}-IT & Instruction & 48.4 & 46.0 & 48.4 & 28.6 & 42.2 & 36.0 & 41.6 \\
        \codellama{7}-Instruct & Instruction & 47.0 & 39.0 & 45.7 & 26.8 & 38.4 & 30.5 & 37.9 \\
        \codellama{13}-Instruct & Instruction & 50.6 & 45.1 & 47.0 & 29.9 & 37.8 & 26.2 & 39.4 \\
        \codellama{34}-Instruct & Instruction & 48.8 & 48.8 & 48.8 & 26.2 & 42.7 & 32.3 & 41.3 \\
        \codellama{70}-Instruct & Instruction & 67.8 & 61.6 & 70.7 & 51.2 & 60.4 & 41.5 & 58.9 \\
        \octocoder{15} & Instruction & 43.9 & 39.0 & 39.6 & 30.5 & 36.0 & 25.0 & 35.7 \\
        \midrule
        Granite-3b-Code-Instruct & Instruction & 51.2 & 43.9 & 41.5 & 31.7 & 40.2 & 29.3 & 39.6 \\
        Granite-8b-Code-Instruct & Instruction & 57.9 & 52.4 & 58.5 & 43.3 & 48.2 & 37.2 & 49.6 \\
        Granite-20B-Code-Instruct & Instruction & 60.4 & 53.7 & 58.5 & 42.1 & 45.7 & 42.7 & 50.5 \\
        Granite-34B-Code-Instruct & Instruction & 62.2 & 56.7 & 62.8 & 47.6 & 57.9 & 41.5 & 54.8 \\    
        \midrule
        \gemma{2}-IT & Instruction & 17.7 & 13.4 & 10.4 & 7.3 & 17.7 & 2.4 & 11.5 \\
        \gemma{7}-IT & Instruction & 28.7 & 17.1 & 29.9 & 18.3 & 18.9 & 6.1 & 19.8\\
        \mistral{7}-Instruct-v0.2 & Instruction & 39.6 & 32.9 & 36.6 & 22.0 & 33.5 & 19.5 & 30.7 \\
        Mixtral-8x7B-Instruct-v0.1 & Instruction & 52.4 & 53.0 & 56.1 & 38.4 & 54.9 & 35.4 & 48.4 \\
        Mixtral-8x22B-Instruct-v0.1 & Instruction & 70.7 & 69.5 & 75.6 & 55.5 & 69.5 & 48.2 & 64.8 \\
        \llamathree{8}-Instruct & Instruction & 60.4 & 30.5 & 30.5 & 22.6 & 46.3 & 31.1 & 36.9 \\
        \llamathree{70}-Instruct & Instruction & 76.2 & 69.5 & 76.2 & 51.8 & 65.2 & 54.3 & 65.5 \\
        \bottomrule \\
        \multicolumn{9}{l}{$^\star$ We could not produce reasonable results for Python generation despite many attempts using}  \\
        \multicolumn{9}{l}{\hspace{1.8mm} different prompts, generation parameters, precision settings.} 
    \end{tabular}}
\end{table}

\subsubsection{HumanEvalSynthesize: Multilingual Code Generation in 6 Languages} 

While most of the prior code LLMs evaluate code generation capabilities only on Python using HumanEval~\citep{human-eval}, we adopt the challenging HumanEvalSynthesize~\citep{octopack} benchmark in our study, which extends Python problems of Humaneval Benchmark to five additional commonly used programming
languages, namely JavaScript, Java, Go, C++, Rust. We evaluate all models in a zero-shot manner using greedy decoding with completion format for the base models, and instruction template for the instruction-tuned models. In constructing prompts for instruction-tuned models, we adhere to the formats provided
in their official examples. We search for a suitable prompt format in the HuggingFace model card,
GitHub repository, and formal publications or technical reports.

Table~\ref{tab:hevalsynthesize} shows the results on of base and instruct models HumanEvalSynthesize benchmark. Granite-3B-Code-Base is the best performing small model with +3\% improvement over CodeGemma-2B. Overall, among base models, \lm achieve the best average performance at 7B-8B scale, 2nd best average performance in 13B-20B size models, and is very close to the best model (falls behind StarCoder2-15B by 0.1\%). While CodeLlama-34B achieves better score on HumanEval Python, Granite-34B-Code-Base achieves much better performance on other languages, leading to a 4\% improvement on average across 6 languages.
Among the instruct models, \lm consistently outperform equivalent size CodeLlama; the 3B, 8B, and 20B models even outperform CodeLlama models that are two times larger.
It's worth noting that even our smaller model, Granite-3B-Code-Instruct, surpasses the performance of CodeLlama-34B-Instruct.
Further, we can also see that \lm outperform much larger state-of-the-art open-source general-purpose language models, including Gemma, Mixtral, and Llama 3 series models.
This shows that domain-specific code models could achieve better performance and efficiency, thus making them more suitable for cost- and performance-sensitive enterprise environments.

\subsubsection{MultiPL-E: Multilingual Code Generation in 18 Languages}

MultiPL-E~\citep{cassano2023multipl} is a canonical benchmark for evaluating code models on a more diverse set of 18 different programming languages. On MultiPL-E, we compare all the base models on 18 languages, by sampling 50 completions per prompt at temperature 0.2 with top-p 0.95, as in~\citep{lozhkov2024starcoder}. Table~\ref{tab:multiple} shows the results of different models on MultiPL-E. As can be seen from the table, there is no single model that works best at every language across all model sizes. In comparison to the similarly sized open-source model CodeLlama-7B, Granite-8B-Code-Base performs the best on 16/18 programming languages. Of the medium models, StarCoder2-15B performs best. Among the large models, Granite-34B-Code-Base does better than CodeLlama-34B on most languages, demonstrating its effectiveness in code generation across a diverse set of languages. 

\begin{table}[t]
    \caption{Pass@1 results on MultiPL-E averaged over 50 samples for each problem. All models are evaluated at temperature 0.2 and top-p 0.95. 
    }
    \label{tab:multiple}
    \centering
    \resizebox{\linewidth}{!}{
    \begin{tabular}{c@{\;}ccccccccc}
\toprule
Model & 
\textbf{C++} & \textbf{C\#} & \textbf{D} & \textbf{Go} & \textbf{Java} & \textbf{Julia} & \textbf{JavaScript} & \textbf{Lua} & \textbf{PHP} \\
\midrule
\starcoderbase{3} & 19.9 & 13.0 & 12.3 & 13.3 & 15.0 & 16.6 & 16.7 & 16.8 & 17.1 \\
\stablecode{3} & 31.5 & 15.0 & 12.3 & 19.7 & 23.9 & 24.9 & 26.0 & 23.7 & 23.8  \\
\starcodertwo{3} & \textbf{32.8} & \textbf{23.4} & \textbf{24.5} & \textbf{24.2} & \textbf{27.0} & \textbf{27.5} & \textbf{29.3} & \textbf{27.6} & \textbf{28.0}  \\
\codegemma{2} & 29.4 & 18.6 & 16.8 & 22.0 & 19.4 & 12.5 & 15.6 & 11.9 & 12.8  \\
Granite-3B-Code-Base & 31.6 & 21.5 & 22.8 & 22.7 & 25.7 & 27.0 & 27.2 & 26.8 & 27.1 \\
\midrule
\starcoderbase{7} & 24.0 & 19.6 & 16.3 & 19.8 & 19.5 & 21.7 & 21.6 & 21.9 & 22.1 \\
\codellama{7} & 29.0 & 21.6 & 20.5 & 21.2 & 24.4 & 26.2 & 26.2 & 26.9 & 26.7 \\
\starcodertwo{7} & 39.1 & 24.7 & 27.6 & 22.2 & 30.5 & 29.6 & 31.8 & 29.8 & 30.2 \\
\codegemma{7} & 43.7 & \textbf{28.2} & 27.6 & 27.9 & 31.3 & \textbf{34.4} & 35.2 & \textbf{35.3} & \textbf{35.7} \\
Granite-8B-Code-Base & \textbf{44.3} & 21.5 & \textbf{30.2} & \textbf{28.0} & \textbf{33.1} & 33.7 & \textbf{35.5} & 33.4 & 33.8 \\
\midrule
\starcoderbase{15}  & 30.2 & 20.6 & 20.4 & 22.0 & 22.9 & 24.6 & 25.2 & 24.9 & 25.2 \\
\codellama{13} & 38.8 & 24.5 & 27.3 & 26.7 & 30.4 & 31.7 & 32.5 & 31.7 & 31.8 \\
\starcodertwo{15} & \textbf{47.4} & \textbf{31.7} & \textbf{35.4} & 27.3 & \textbf{36.6} & \textbf{37.5} & 38.5 & \textbf{38.5} & 31.9 \\
Granite-20B-Code-Base & 43.3 & 30.6 & 16.2 & \textbf{29.4} & 35.9 & 31.8 & \textbf{38.9} & 30.5 & \textbf{38.3} \\
\midrule
\codellama{34} & 45.9 & 31.0 & \textbf{30.5} & 28.4 & \textbf{35.2} & \textbf{36.1} & 37.0 & 36.4 & 37.1 \\
Granite-34B-Code-Base & \textbf{46.7} & \textbf{32.8} & 18.7 & \textbf{33.2} & 33.3 & 31.8 & \textbf{44.5} & \textbf{38.2} & \textbf{42.5}  \\
\midrule
\midrule
Model & \textbf{Perl} & \textbf{R} & \textbf{Ruby} & \textbf{Racket} & \textbf{Rust} & \textbf{Scala} & \textbf{Bash} & \textbf{Swift} & \textbf{TypeScript} \\
\midrule
\starcoderbase{3} & 11.0 & 16.2 & 14.2 & 15.3 & 14.4 & 16.4 & 4.2 & 13.2 & 22.4 \\
\stablecode{3} & 9.7 & 22.5 & 18.7 & 20.7 & 19.1 & 14.6 & 8.4 & 17.5 & 29.5  \\
\starcodertwo{3} & 13.2 & \textbf{26.7} & \textbf{25.2} & \textbf{24.6} & \textbf{25.3} & 21.1 & \textbf{12.5} & \textbf{23.2} &  \textbf{35.5} \\
\codegemma{2} & 2.5 & 11.3 & 9.9 & 10.9 & 11.6 & 25.6 & 1.6 & 10.7 & 1.3  \\
Granite-3B-Code-Base & \textbf{18.8} & 25.8 & 24.1 & 24.1 & 24.0 & \textbf{26.9} & 7.0 & 22.0 & 31.6 \\
\midrule
\starcoderbase{7} & 16.8 & 21.1 & 19.9 & 20.0 & 20.1 & 21.2 & 7.5 & 18.5 & 27.6  \\
\codellama{7} & 17.4 & 25.7 & 24.7 & 24.2 & 24.9 & 25.5 & 10.0 & 22.8 & 33.7  \\
\starcodertwo{7} & 17.6 & 29.0 & 27.3 & 27.2 & 27.6 & 22.0 & 13.2 & 25.3 & 37.0  \\
\codegemma{7} & \textbf{31.2} & \textbf{34.0} & \textbf{33.1} & \textbf{32.0} & \textbf{33.6} & \textbf{39.0} & 11.0 & \textbf{30.8} & \textbf{45.2}  \\
Granite-8B-Code-Base & 18.8 & 32.2 & 30.4 & 30.4 & 30.4 & 26.9 & \textbf{13.6} & 27.9 & 31.6 \\
\midrule
\starcoderbase{15} & 16.8 & 23.9 & 22.0 & 22.6 & 22.3 & 28.6 & 11.2 & 20.4 & 32.3  \\
\codellama{13} & 22.5 & 30.1 & 28.8 & 28.4 & 29.0 & 29.7 & 13.8 & 26.6 & 40.6 \\
\starcodertwo{15} & \textbf{36.6} & \textbf{37.2} & \textbf{36.4} & \textbf{35.9} & \textbf{36.6} & 18.7 & \textbf{18.7}  & \textbf{33.5} & \textbf{43.9}  \\
Granite-20B-Code-Base & 26.5 & 15.5 & 28.1 & 17.9 & 35.7 & \textbf{37.5} & 16.7 & 27.6 & 38.7 \\
\midrule
\codellama{34} & 28.9 & \textbf{35.4} & 33.8 & \textbf{33.4} & 34.3 & 32.9 & 16.2 & 31.5 & 39.6 \\
Granite-34B-Code-Base & \textbf{31.5} & 25.4 & \textbf{33.9} & 18.2 & \textbf{38.9} & \textbf{41.7} & \textbf{19.3} & \textbf{36.5} & \textbf{41.5} \\
\bottomrule
\end{tabular}} 
\end{table}

\begin{table}
\parbox{.45\linewidth}{
\caption{Pass@1 on MBPP and MBPP+. Results in the table represent zero-shot evaluation using greedy decoding.}
\label{tab:mbpp}
\centering
\resizebox{\linewidth}{!}
{\begin{tabular}{c cc}
    \toprule
    \textbf{Model} & \textbf{MBPP} & \textbf{MBPP+}\\
    \midrule
    \starcoderbase{3} & 29.4 & 37.8 \\
    \stablecode{3}   & 34.8 & 43.3 \\
    \starcodertwo{3}  & \textbf{42.4} & \textbf{48.6} \\ 
    \codegemma{2} & 30.4 & 30.8  \\
    Granite-3B-Code-Base  & 36.0 & 45.1 \\
    \midrule                              
    \starcoderbase{7}  & 34.8 & 42.1 \\
    \codellama{7}  & 39.0 & 42.3 \\
    \starcodertwo{7}  & 45.4 & 46.7 \\
    \codegemma{7}  & \textbf{53.0} & \textbf{54.9} \\
    Granite-8B-Code-Base  & 42.2 & 49.6 \\
    \midrule                              
    \starcoderbase{15} & 37.4 & 46.1 \\
    \codellama{13} & 30.6 & 30.1 \\
    \starcodertwo{15} & \textbf{51.2} & \textbf{56.6} \\
    Granite-20B-Code-Base  & 43.8 & 51.6 \\
    \midrule                              
    \codellama{34}  & \textbf{48.6} & \textbf{53.6} \\
    Granite-34B-Code-Base & 47.2 & 53.1 \\
\bottomrule
\end{tabular}}}
\hfill
\parbox{.45\linewidth}{
\caption{Average exact match (EM) and edit similarity (ES) on RepoBench v1.1.  All models are evaluated at temperature 0.2 and top-p 0.95.}
\label{tab:repobench}
\centering
\resizebox{\linewidth}{!}
   {\begin{tabular}{c cc cc}
    \toprule
    \multirow{2}*{\textbf{Model}} &
    \multicolumn{2}{c}{\textbf{Python}} &
    \multicolumn{2}{c}{\textbf{Java}}
    \\ \cmidrule(lr){2-3} \cmidrule(lr){4-5}
    & \textbf{EM} & \textbf{ES} 
    & \textbf{EM} & \textbf{ES}  \\
    \midrule
    \starcoderbase{3} & \textbf{29.9} &	\textbf{69.3} & \textbf{36.0} &	\textbf{74.1}\\
    \stablecode{3} & 29.4 & 68.5 & 34.9 & 72.8\\
    \starcodertwo{3} & 27.2 & 67.0 & 35.9 & 74.1 \\
    \codegemma{2} & 26.2 & 66.9 & 33.6 & 71.6 \\
    Granite-3B-Code-Base & 27.1 &	66.8 &	34.9 &73.9\\

    \midrule
    \starcoderbase{7}  & 27.1 & 66.5 & 36.5 & 75.0 \\
    \codellama{7} &  29.2 & 67.4 & 37.9 & \textbf{76.6}\\ 
    \starcodertwo{7} & 28.1 & 67.6 & 37.0 & 75.2 \\
    \codegemma{7} & \textbf{36.8} & \textbf{72.7} & 38.3 & 74.3 \\
    Granite-8B-Code-Base & 31.8	& 69.5	& \textbf{38.4} &76.4 \\
    \midrule
    \starcoderbase{15}  & 29.4 &	67.8 &37.1 &75.4 \\
    \codellama{13}   & 31.4 & 68.8 & 39.4 & 77.4 \\
    \starcodertwo{15}  & 31.3 &69.6 & 39.9 &	77.3\\
    Granite-20B-Code-Base & \textbf{38.0}& \textbf{72.2} & \textbf{42.3} & \textbf{78.1} \\
    \midrule
    \codellama{34} & 34.4 & 70.2 & 40.8 & \textbf{78.4} \\
    Granite-34B-Code-Base & \textbf{35.9} & \textbf{71.5} & \textbf{42.0}	&77.7\\
    \bottomrule
\end{tabular}}
}
\end{table}

\subsubsection{MBPP and MBPP+: Code Generation in Python}

MBPP~\citep{mbpp} and MBPP+~\citep{evalplus} are two of the most widely studied benchmarks for evaluating code models. While the prompt for each MBPP problem includes a natural language description followed by a few tests, MBPP+ consists of 35$\times$ more tests than the original benchmarks. We use greedy decoding and report the mean pass@1 for all the models. Table~\ref{tab:mbpp} summarizes the results of different base models. As we can see, Granite-3B-Code-Base significantly outperforms CodeGemma-2B but falls short of StarCoder2-3B on both benchmarks. At mid parameter ranges, Granite Code models beat both CodeLlama-7B and CodeLLama-13B by a margin of $\sim$5\% and $\sim$15\% on average respectively. Additionally, Granite-34B-Code-Base is very competitive with CodeLlama-34B with only a difference of 0.9\% on average across both benchmarks.

\subsubsection{DS1000: Data Science Tasks in Python}
DS-1000~\citep{lai2023ds} is a widely studied benchmark which offers a comprehensive collection of 1,000 data science workflows across seven different libraries, from Matplotlib to TensorFlow. We use temperature 0.2 and top-p 0.95 to generate 40 samples per each library and report mean pass@1 with code completion setting for all the models up to 8B parameters. Results on DS-1000 are summarized in Table~\ref{tab:ds1000}.  Of the small models, StarCoder2-3B performs the best. Granite-3B-Code-Base is in second place, outperforming CodeGemma-2B by more than 12 points on average across 7 libraries. Granite-8B-Code-Base achieves the best average performance of 34.5\% outperforming all other models of the similar parameter sizes.

The Granite Code models achieve relatively high accuracy across all sizes (e.g., outperforming CodeGemma at 2B-3B scale, StarCoder2 at 7B-8B scale and CodeLlama models with half of the sizes). This shows that our
Granite Code models are not only capable of generating good code but also of using libraries more accurately in real data science workflows. 

\subsubsection{RepoBench, CrossCodeEval: Repository-Level Code Generation}

Code generation in practice often occurs within the context of a repository rather than in isolated files. Thus, we use RepoBench~\citep{liu2023repobench} and CrossCodeEval~\citep{ding2024crosscodeeval}, to evaluate repository-level code completion capabilities of different models.

On RepoBench, we evaluate using level 2k across three settings: cross-file-first (12,000 data points),
cross-file-random (5,000 data points), and in-file (7,000 data points). We report the average edit similarity and exact match across the settings. Following \cite{liu2023repobench},
we set generation temperature to 0.2 and the top-p sampling parameter to 0.95 for all models. We constrain the models to generate a maximum of 64 new tokens per prompt, and the first non-empty and non-comment
line of the output was selected as the prediction. 

\begin{table}[t]
    \caption{Mean pass@1 accuracy averaged over 40 samples on DS-1000. All models are evaluated at temperature $0.2$ and top-p $0.95$.
    } 
    \label{tab:ds1000}
    \centering
    \resizebox{\linewidth}{!}{
    \begin{tabular}{lcccccccc}
    \toprule\\[-6pt]
    \textbf{Model} & \textbf{\small Matplotlib} & \textbf{\small NumPy} & \textbf{\small Pandas} & \textbf{\small PyTorch} & \textbf{\small SciPy} & \textbf{\small Scikit-Learn} & \textbf{\small TensorFlow} & \textbf{Avg} \\
    \midrule

    \starcoderbase{3} & 32.1 & 16.8 & 5.3 & 9.2 & 13.2 & 10.5 & 17.2 & 14.2 \\
    \stablecode{3} & 42.5 & 24.5 & \textbf{16.2} & 15.4 & 13.5 & 20.2 & 27.7 & 22.7 \\ 
    \starcodertwo{3} & 45.5 & \textbf{27.7} & \textbf{16.2} & 12.9 & 15.8 & \textbf{30.8} & 22.8 & \textbf{25.0} \\
    \codegemma{2} & 30.3 & 17.7 & 5.5 & 4.4 & 10.3 & 2.6 & 4.4 & 10.7 \\
    Granite-3B-Code-Base & \textbf{43.3} & \textbf{27.7} & 11.0 & \textbf{19.1} & \textbf{21.7} & 16.5 & \textbf{24.4} & 23.4 \\
    \midrule
    \starcoderbase{7} & 38.0 & 23.0 & 8.2 & 13.1 & 13.7 & 24.5 & 14.6 & 19.1 \\
    \codellama{7} & 46.3 & 21.6 & 13.9 & 12.2 & 17.5 & 16.7 & 20.6 & 21.5 \\  
    \starcodertwo{7} & 53.6 & 33.3 & 16.9 & 16.2 & 20.6 & 22.2 & 31.9 & 27.8 \\
    \codegemma{7} & \textbf{56.1} & 37.2 & 20.9 & 20.5 & 24.5 & \textbf{34.7} & 31.1 & 32.1 \\ 
    Granite-8B-Code-Base & 51.6 & \textbf{40.0} & \textbf{23.4} & \textbf{32.4} & \textbf{22.6} & 29.6 & \textbf{42.2} & \textbf{34.5} \\
    \bottomrule
    \end{tabular}
    }
\end{table}

\begin{table}[!t]

\centering
\caption{CrossCodeEval \citep{ding2023crosscodeeval} evaluation results. We report Code Match (Edit Similarity) and Identifier Match (F1) results for four languages.
}
\label{tab:crosscodeeval}
    \resizebox{1.0\linewidth}{!}{
\begin{tabular}{ccccccccc}
\toprule
\multirow{3}{*}{\textbf{Model}} & \multicolumn{2}{c}{\textbf{Python}}                           & \multicolumn{2}{c}{\textbf{Java}}                             & \multicolumn{2}{c}{\textbf{TypeScript}}                       & \multicolumn{2}{c}{\textbf{C\#}}                              \\\cmidrule(lr){2-3} \cmidrule(lr){4-5}  \cmidrule(lr){6-7} \cmidrule(lr){8-9}
                       & \textbf{Code ES} & \textbf{ID F1} & \textbf{Code ES} & \textbf{ID F1}& \textbf{Code ES} & \textbf{ID F1}& \textbf{Code ES} & \textbf{ID F1} \\
                       \midrule
\starcoderbase{3} & 63.5	&53.8	&63.3	&55.7	&44.2	&40.8&	65.3&	45.0        \\
\stablecode{3} & 65.3&	56.1&	\textbf{68.2} &	\textbf{61.0} & \textbf{60.9} &	\textbf{55.7} &	59.9&	41.7       \\
\starcodertwo{3} & \textbf{65.5}	& \textbf{56.7} &	64.8&	57.3&	44.7&	41.3& \textbf{66.0} & \textbf{47.5}   \\
\codegemma{2} & 60.5&	50.6&	55.1&	46.4&	55.6&	49.0&	44.2&	27.9\\
Granite-3B-Code-Base & 65.1&	56.0	&64.1	&56.6	&43.2&	39.4&	65.9&	46.6\\
\midrule
\starcoderbase{7} & 65.3&	56.3&	65.4&	58.0&	46.2&	43.2	&66.1&	47.2\\
\codellama{7}  & 64.9	&55.4&	65.0&	57.8&	62.1	&56.9&	65.1	&46.9    \\
\starcodertwo{7}   & 66.5	&57.5	& \textbf{67.0}	& \textbf{59.8}	&46.9	&43.3	& \textbf{67.2} &	48.6        \\
\codegemma{7} & \textbf{68.1}&	\textbf{59.3} &	65.9	&59.6&	\textbf{63.5}	& \textbf{58.5} &56.2	&42.1\\
Granite-8B-Code-Base & 66.3	&57.8	&66.5&	59.0&	45.1	&42.0	&66.6	& \textbf{48.7} \\
\midrule
\starcoderbase{15}  & 66.0 &	57.3&	67.0&	60.2&	46.6	&43.3	&66.4	&47.7    \\
\codellama{13} & 66.2	&57.4	&66.8&	59.5&	63.5	& \textbf{58.6} &	65.6	&48.3     \\
\starcodertwo{15} & 68.1&	\textbf{59.7}	&68.1	&\textbf{61.5} &	47.0	&43.7& \textbf{68.5} & \textbf{51.0}    \\
Granite-20B-Code-Base & \textbf{68.2}	&58.1& \textbf{68.4} &60.4	&\textbf{48.3}	&43.1	&67.5	&48.4\\
\midrule
\codellama{34}  & \textbf{69.3}&	\textbf{59.6}&	68.2	&\textbf{61.1}&	\textbf{64.4}&	\textbf{56.9}&	67.2&	\textbf{49.6}  \\
Granite-34B-Code-Base & 68.3 &	58.5&	\textbf{68.6}	&60.8&	48.9	&43.6&	\textbf{67.5}&	49.1\\
\bottomrule
\end{tabular}
}
\end{table}

For CrossCodeEval, following \cite{ding2024crosscodeeval}, we use a max sequence length of 2k using the retrieve-and-generate (RG) method with OpenAI’s ada embedding.  We set the maximum cross-file context to 512 tokens and the max generation token to 50 tokens for all the models. We use the uniform prompt formatting in the original implementation, with a temperature of 0.2 and top-p of 0.95 for all model generations.  
Max sequence length was set to 8,192 for all models, with the exception of Granite-3B-Code-Base (2,048) and Granite-8B-Code-Base (4,096), given their respective context lengths.

Table~\ref{tab:repobench} shows the performance of different models on RepoBench v1.1. Granite-3B-Code-Base demonstrates notable performance among the smaller models, with StarCoderBase-3B achieving the leading performance metrics. Among the medium models, Granite-8B-Code-Base shows very strong performance on Java, while ranks second best one in Python, with CodeGemma-7B being the best performing on both metrics. Among larger models, Granite-20B-Code not only outperforms StarCoder2-15B but also CodeLlama-34B on all 4 metrics across both programming languages. This demonstrates strong repository-level code generation capabilities of the Granite Code models despite being not trained with repo-level file packing as in~\citep{lozhkov2024starcoder,codegemma}; we leave this as an interesting future work to further improve performance of our models. 

Results on CrossCodeEval are shown in Table~\ref{tab:crosscodeeval}. As can be seen from the table, among the similar sized models, CodeGemma-7B is best on Python and TypeScript, while StarCoder2-7B performs best on Java and C\#. Likewise, Granite-20B-Code-Base outperforms CodeLlama-13B on 3 programming languages (Python, Java, C\#), while falls behind on TypeScript. Across all model sizes and programming languages, there is no single model that is best at all the metrics, similar to the findings in MultiPL-E. This indicates that achieving uniformly high performance across all programming languages remains challenging.

\subsubsection{FIM: Infilling Evaluations}

\begin{table}[t]
    \caption{Exact-match on FIM-task  \citep{allal2023santacoder}. All models are evaluated using greedy decoding with maximum new tokens set to 512.}
        \label{tab:fim}
    \centering
    \small
    \begin{tabular}{cccc|c}
    \toprule
        \textbf{Model} & \textbf{Java} & \textbf{JavaScript} & \textbf{Python} & \textbf{Avg.} \\ \midrule
        \starcoderbase{3} & 76.0 & 68.5 & 53.6 & 66.0 \\
    \stablecode{3} & 64.2 & \textbf{74.5} & 59.6 & 66.1 \\
    \starcodertwo{3} & 76.0 & 73.5 & 59.4 & 69.6\\
    Granite-3B-Code-Base & \textbf{79.7} & 71.6 & \textbf{61.8} & \textbf{71.0} \\
    \midrule
    \starcoderbase{7} & 81.1 & 74.5 & 62.0 & 72.5 \\
    \starcodertwo{7} & 82.1 & 78.4 & 61.5 & 74.0 \\
    Granite-8B-Code-Base & \textbf{83.6} & \textbf{79.9} & \textbf{66.3} & \textbf{76.6} \\
    \midrule
    \starcoderbase{15}  & 74.6 & 74.6 & 63.1 & 70.8 \\
    \starcodertwo{15} & 61.1 & 54.8 & 48.4 & 54.8 \\
    Granite-20B-Code-Base & 79.4 & \textbf{82.2} & \textbf{66.8} & \textbf{76.1} \\
    Granite-34B-Code-Base & \textbf{80.8} & 79.4 & 67.9 & 76.0 \\
        \bottomrule
    \end{tabular}
\end{table}

Granite Code models are trained for code
completion purposes using FIM objective, as described in Sec.~\ref{sec:loss}. We use SantaCoder-FIM benchmark~\citep{allal2023santacoder}, for infilling evaluations which tests the ability of models to fill in a single line of code in Python,
JavaScript, and Java solutions to HumanEval. We use greedy decoding and report the mean exact match for all the models. 
Table~\ref{tab:fim} shows that Granite Code models significantly outperforms StarCoder and StarCoder2 across all model sizes, demonstrating it to be excellent well-rounded models for code completion use cases. Moreover, we observe no performance improvement in scaling the model sizes from 8B to 34B, indicating that smaller models are often more suitable for FIM code completion tasks.

\subsection{Code Explanation and Fixing}

While most of the prior code LLMs primarily focus on evaluating performance using code generation benchmarks, users may want to use these models in other challenging scenarios beyond synthesis like explaining and fixing codes. 
Thus, following~\citep{octopack}, we test the performance of different code models on the code explanation and fixing variants of HumanEvalPack benchmark, spanning 6 different programming languages. 
For both HumanEvalExplain and HumanEvalFix, we evaluate all models in a zero-shot manner using greedy decoding with completion format for the base models, and instruction template for the instruction-tuned models. 

The results of the HumanEvalExplain benchmark are shown in Table~\ref{tab:hevalexplain}. 
Granite Code Base models significantly outperform other SOTA base code LLMs, including StarCoder2 and CodeGemma, by a large margin. 
Interestingly, Granite-8B-Code-Base beats CodeLlama-34B by 9.3\% on average, while being close to CodeLlama-70B. 
We attribute this performance to our data mixture and base model training decisions. 
After instruction tuning, performance of all the base models significantly improves across languages. 
Among code instruct models, Granite-34B-Code-Instruct performs the best reaching the average score of 41.9\%, which is very close of 41.1\% score of CodeLlama-70B-Instruct. 
Remarkably, CodeGemma-7B-IT gains the most improvement from instruction tuning but still falls behind Granite-8b-Code-Instruct by 2.5\% on average. 
Mixtral-8x22B-Instruct-v0.1 performs best among all models benchmarked with a significant
margin, indicating the potential advantage of bigger models and training on general natural language data could potentially help on this task.

\begin{figure}
\begin{center}
\includegraphics[width=1\linewidth]{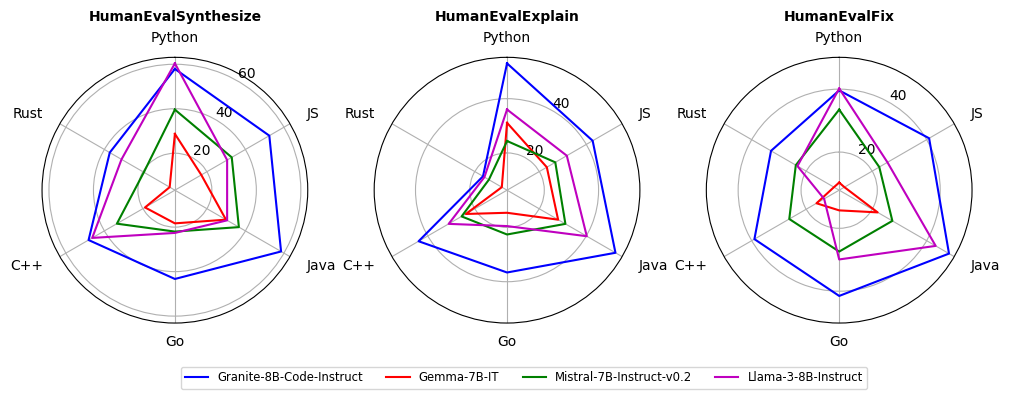}
\end{center} 
\caption{Performance of Granite-8B-Code-Instruct, Mistral-7B-Instruct-v0.2, Gemma-7B-IT, and Llama-3-8B-Instruct on HumanEvalPack. Best viewed in color.}
\label{fig:granite-llama}
\end{figure}

Table~\ref{tab:hevalfix} reports the results on HumanEvalFix. 
Like HumanEvalExplain, \lm base models significantly outperform other base models. 
Notably, Granite-8B-Code-Base again shows impressive performance, making it close to CodeLlama-70B and Llama-3-70B. 
After instruction tuning, we observe a performance improvement on almost all models.
Notably, our 8B and 20B instruct models achieve the best performance among models with less that has less than 34B parameters.
However, we see a significant performance improvement (about 10 points) after moving to larger models with more than 34B parameters. 
Among large instruct models, Granite-34B-Code-Instruct performs similarly to other models with at least twice the parameters, thus having a better cost and performance balance.

Figure~\ref{fig:granite-llama} compares the performance of Granite-8B-Code-Instruct with state-of-the-art open-source instruction-tuned general LLMs. Granite-8B-Code-Instruct consistently outperforms the compared models, emphasizing the need for domain-specific code models. To summarize, these results show that both our base and instruct models are capable of generating good code but also in code fixing and explanation, demonstrating their ability to solve diverse coding tasks in enterprise software development. 

\begin{table}
    \caption{Pass@1 performance on HumanEvalExplain.}
    \label{tab:hevalexplain}
    \centering
    \small
    \resizebox{\linewidth}{!}
    {\begin{tabular}{cc|cccccc|c}
        \toprule
        \textbf{Model} & \textbf{Prompt} & \textbf{Python} & \textbf{JavaScript} & \textbf{Java} & \textbf{Go} & \textbf{C++} & \textbf{Rust} & \textbf{Avg.} \\
        \midrule
        \multicolumn{9}{c}{Base Models} \\
        \midrule
        \starcoderbase{3} & Completion & 11.0 & 10.4 & 14.6 & 11.0 & 13.4 & 11.0 & 11.9 \\
        \stablecode{3} & Completion & 11.0 & 7.9 & 22.0 & 4.3 & 14.6 & 14.0 & 12.3 \\
        \starcodertwo{3} & Completion & 12.2 & 13.4 & 19.5 & 6.7 & 14.0 & 12.8 & 13.1 \\
        \codegemma{2} & Completion & 20.7 & 15.9 & 20.7 & 12.8 & 17.7 & 15.9 & 17.3 \\
        Granite-3B-Code-Base & Completion & \textbf{25.0} & \textbf{18.9} & \textbf{29.9} & \textbf{17.1} & \textbf{26.8} & \textbf{14.0} & \textbf{21.9} \\
        \midrule
        \starcoderbase{7} & Completion & 14.0 & 17.1 & 17.7 & 10.4 & 17.1 & 12.8 & 14.8 \\
        \codellama{7} & Completion & 11.0 & 14.0 & 16.5 & 9.8 & 17.7 & 14.6 & 13.9 \\
        \starcodertwo{7} & Completion & 4.9 & 12.8 & 22.0 & 4.9 & 22.0 & 14.6 & 13.5 \\
        \codegemma{7} & Completion & 13.1 & 13.8 & 2.0 & 8.0 & 18.6 & 18.9 & 12.4 \\
        Granite-8B-Code-Base & Completion & \textbf{23.5} & \textbf{32.3} & \textbf{25.0} & \textbf{23.2} & \textbf{28.0} & \textbf{19.5} & \textbf{26.4} \\
        \midrule
        \starcoderbase{15} & Completion & 9.8 & 15.2 & \textbf{24.4} & 9.1 & 20.1 & 13.4 & 15.3 \\
        \codellama{13} & Completion & 13.4 & 14.0 & 23.2 & 9.8 & 15.9 & 13.4 & 15.0 \\
        \starcodertwo{15} & Completion & \textbf{20.1} & \textbf{19.5} & 7.3 & 9.8 & 23.8 & \textbf{21.3} & 17.0 \\
        Granite-20B-Code-Base & Completion & 17.1 & 18.3 & 23.2 & \textbf{10.4} & \textbf{25.6} & 18.3 & \textbf{18.8} \\
        \midrule
        \codellama{34} & Completion & 11.6 & 18.3 & 22.0 & 9.8 & 20.1 & 20.7 & 17.1 \\
        Granite-34B-Code-Base & Completion & \textbf{42.7} & \textbf{26.2} & \textbf{47.0} & \textbf{26.8} & \textbf{36.6} & \textbf{25.0} & \textbf{34.1} \\
        \codellama{70} & Completion & 24.4 & 30.5 & 43.9 & 19.5 & 31.1 & 20.1 & 28.2 \\
        \midrule
        \gemma{2} & Completion & 9.8 & 9.8 & 14.6 & 7.9 & 14.0 & 9.1 & 10.9 \\
        \gemma{7} & Completion & 10.4 & 18.3 & 19.5 & 9.8 & 18.3 & 14.0 & 15.0 \\
        \mistral{7}-v0.2 & Completion & 22.0 & 23.8 & 34.8 & 16.5 & 14.6 & 12.8 & 20.7 \\
        Mixtral-8x7B-v0.1 & Completion & 17.1 & 18.3 & 35.4 & 19.5 & 18.9 & 15.2 & 20.7 \\
        Mixtral-8x22B-v0.1 & Completion & 29.9 & 20.1 & 40.2 & 17.7 & 22.0 & 17.7 & 24.6 \\
        \llamathree{8} & Completion & 15.2 & 14.0 & 18.9 & 5.5 & 18.3 & 8.5 & 13.4 \\
        \llamathree{70} & Completion & 12.2 & 18.9 & 20.7 & 9.1 & 16.5 & 15.9 & 15.6 \\
        \midrule
        \multicolumn{9}{c}{Instruct Models} \\
        \midrule
        \codegemma{7}-IT & Instruction & 48.2 & 40.9 & 51.8 & 31.1 & 33.5 & 25.0 & 38.4 \\
        \codellama{7}-Instruct & Instruction & 29.9 & 29.9 & 32.9 & 19.5 & 25.0 & 13.4 & 25.1 \\
        \codellama{13}-Instruct & Instruction & 38.4 & 28.0 & 36.0 & 22.6 & 26.8 & 14.6 & 27.7 \\
        \codellama{34}-Instruct & Instruction & 42.1 & 31.1 & 39.6 & 20.1 & 31.7 & 17.1 & 30.3 \\
        \codellama{70}-Instruct  & Instruction & 47.0 & 40.2 & 54.9 & 34.1 & 46.3 & 23.8 & 41.1 \\
        \octocoder{15} & Instruction & 37.8 & 26.8 & 31.1 & 18.3 & 22.6 & 14.0 & 25.1 \\
        \midrule
        Granite-3b-Code-Instruct & Instruction & 39.6 & 26.8 & 39.0 & 14.0 & 23.8 & 12.8 & 26.0 \\
        Granite-8b-Code-Instruct & Instruction & 53.0 & 42.7 & 52.4 & 36.6 & 43.9 & 16.5 & 40.9 \\
        Granite-20B-Code-Instruct & Instruction & 44.5 & 42.7 & 49.4 & 32.3 & 42.1 & 18.3 & 38.2 \\
        Granite-34B-Code-Instruct & Instruction & 53.0 & 45.1 & 50.6 & 36.0 & 42.7 & 23.8 & 41.9\\
        \midrule
        \gemma{2}-IT & Instruction & 9.8 & 12.8 & 13.4 & 8.5 & 13.4 & 3.0 & 10.2 \\
        \gemma{7}-IT & Instruction & 31.1 & 23.2 & 28.0 & 14.6 & 23.8 & 8.5 & 21.5 \\
        \mistral{7}-Instruct-v0.2 & Instruction & 24.4 & 26.8 & 31.1 & 22.6 & 25.6 & 14.0 & 24.1\\
        Mixtral-8x7B-Instruct-v0.1 & Instruction & 47.0 & 40.9 & 48.2 & 28.0 & 32.9 & 25.0 & 37.0 \\
        Mixtral-8x22B-Instruct-v0.1 & Instruction & 67.1 & 56.7 & 67.7 & 44.5 & 64.0 & 39.6 & 56.6 \\
        \llamathree{8}-Instruct & Instruction & 36.0 & 31.7 & 40.2 & 19.5 & 31.1 & 15.9 & 29.1 \\
        \llamathree{70}-Instruct & Instruction & 50.6 & 47.6 & 57.9 & 34.8 & 48.2 & 33.5 & 45.4 \\
        \bottomrule
    \end{tabular}} \vspace{-4mm}
\end{table}
\begin{table}
    \caption{Pass@1 performance on HumanEvalFix.}
    \label{tab:hevalfix}
    \centering
    \small
    \resizebox{\linewidth}{!}
    {\begin{tabular}{cc|cccccc|c}
        \toprule
        \textbf{Model} & \textbf{Prompt} & \textbf{Python} & \textbf{JavaScript} & \textbf{Java} & \textbf{Go} & \textbf{C++} & \textbf{Rust} & \textbf{Avg.} \\
        \midrule
        \multicolumn{9}{c}{Base Models} \\
        \midrule
        \starcoderbase{3} & Completion & 12.2 & 9.8 & 6.1 & 7.9 & 1.8 & 0.6 & 6.4\\
        \stablecode{3} & Completion & 11.0 & 7.3 & 20.1 & 10.4 & 12.8 & 1.2 & 10.5 \\
        \starcodertwo{3} & Completion & \textbf{18.3} & 15.9 & 12.8 & 12.8 & 5.5 & 0.6 & 11.0 \\
        \codegemma{2} & Completion & 4.3 & 7.3 & 3.0 & 9.8 & 1.8 & 0.0 & 4.4 \\
        Granite-3B-Code-Base & Completion & \textbf{18.3} & \textbf{23.2} & \textbf{29.9} & \textbf{24.4} & \textbf{16.5} & \textbf{3.7} & \textbf{19.3} \\
        \midrule
        \starcoderbase{7} & Completion & 15.9 & 21.3 & 17.1 & 14.6 & 5.5 & 0.6 & 12.5 \\
        \codellama{7} & Completion & 15.9 & 14.0 & 23.8 & 15.2 & 5.5 & 11.0 & 14.2 \\
        \starcodertwo{7} & Completion & 5.5 & 13.4 & 15.9 & 11.0 & 7.3 & 0.6 & 8.9 \\
        \codegemma{7} & Completion & 8.5 & 5.5 & 20.1 & 14.6 & 7.9 & 3.7 & 10.1 \\
        Granite-8B-Code-Base & Completion & \textbf{22.6} & \textbf{35.4} & \textbf{38.4} & \textbf{37.2} & \textbf{28.7} & \textbf{15.2} & \textbf{29.6} \\
        \midrule
        \starcoderbase{15} & Completion & 10.4 & 17.7 & 17.1 & 18.9 & 9.8 & 3.7 & 12.9 \\
        \codellama{13} & Completion & 6.1 & 9.1 & 17.1 & 9.8 & 6.1 & 10.4 & 9.5 \\
        \starcodertwo{15} & Completion & 9.1 & 18.9 & 25.0 & 37.2 & 25.0 & 16.5 & 21.9 \\
        Granite-20B-Code-Base & Completion & 23.2 & 23.8 & 14.6 & 26.2 & 15.2 & 3.0 & 17.7 \\
        \midrule
        \codellama{34} & Completion & 14.0 & 20.7 & 20.1 & 26.2 & 31.1 & 6.1 & 19.7 \\
        Granite-34B-Code-Base & Completion &  \textbf{20.1} & \textbf{30.5} & \textbf{40.9} & \textbf{34.1} & \textbf{39.0} & \textbf{12.2} & \textbf{29.5} \\
        \codellama{70} & Completion & 12.8 & 31.1 & 41.5 & 42.1 & 38.4 & 31.1 & 32.8 \\
        \midrule
        \gemma{2} & Completion & 1.2 & 2.4 & 4.3 & 2.4 & 1.2 & 0.0 & 1.9 \\
        \gemma{7} & Completion & 1.8 & 1.2 & 26.2 & 7.3 & 6.7 & 1.2 & 7.4 \\
        \mistral{7}-v0.2 & Completion & 3.0 & 2.4 & 6.1 & 9.1 & 5.5 & 0.6 & 4.5 \\
        Mixtral-8x7B-v0.1 & Completion & 11.0 & 12.8 & 18.3 & 25.0 & 15.9 & 3.0 & 14.3 \\
        Mixtral-8x22B-v0.1 & Completion & 17.1 & 30.5 & 23.8 & 30.5 & 32.3 & 11.0 & 24.2 \\
        \llamathree{8} & Completion & 2.6 & 3.7 & 5.5 & 3.7 & 1.3 & 0.9 & 2.9 \\
        \llamathree{70} & Completion & 31.7 & 33.5 & 39.0 & 28.0 & 28.7 & 12.8 & 28.9 \\
        \midrule
        \multicolumn{9}{c}{Instruct Models} \\
        \midrule
        \codegemma{7}-IT & Instruction & 46.3 & 45.7 & 52.4 & 48.2 & 43.9 & 38.4 & 45.8 \\
        \codellama{7}-Instruct & Instruction & 19.5 & 18.9 & 13.4 & 14.6 & 6.1 & 4.3 & 12.8 \\
        \codellama{13}-Instruct & Instruction & 18.9 & 18.9 & 24.4 & 22.6 & 9.8 & 0.0 & 15.8 \\
        \codellama{34}-Instruct & Instruction & 37.8 & 28.0 & 37.2 & 24.4 & 24.4 & 17.7 & 28.2 \\
        \codellama{70}-Instruct & Instruction & 64.6 & 52.4 & 57.3 & 51.8 & 51.2 & 40.9 & 53.0 \\
        \octocoder{15} & Instruction & 28.0 & 28.7 & 34.1 & 26.8 & 25.0 & 15.9 & 26.4 \\
        \midrule
        Granite-3b-Code-Instruct & Instruction & 26.8 & 28.0 & 33.5 & 27.4 & 31.7 & 16.5 & 27.3 \\
        Granite-8b-Code-Instruct & Instruction & 39.6 & 40.9 & 48.2 & 41.5 & 39.0 & 32.9 & 40.4 \\ 
        Granite-20B-Code-Instruct & Instruction & 43.9 & 43.9 & 45.7 & 41.5 & 41.5 & 29.9 & 41.1 \\
        Granite-34B-Code-Instruct & Instruction & 54.9 & 47.6 & 55.5 & 51.2 & 47.0 & 45.1 & 50.2 \\
        \midrule
        \gemma{2}-IT & Instruction & 18.9 & 15.9 & 25.6 & 13.4 & 15.9 & 10.4 & 16.7 \\
        \gemma{7}-IT & Instruction & 10.3 & 9.8 & 22.0 & 14.3 & 16.2 & 9.8 & 13.7 \\
        \mistral{7}-Instruct-v0.2 & Instruction & 33.5 & 22.6 & 27.4 & 27.4 & 26.2 & 23.8 & 26.8 \\
        Mixtral-8x7B-Instruct-v0.1 & Instruction & 44.5 & 37.8 & 47.6 & 38.4 & 39.0 & 28.7 & 39.3 \\
        Mixtral-8x22B-Instruct-v0.1 & Instruction & 59.1 & 53.7 & 66.5 & 55.5 & 56.1 & 45.7 & 56.1 \\
        \llamathree{8}-Instruct & Instruction & 40.2 & 25.6 & 43.3 & 29.9 & 13.4 & 23.2 & 29.3 \\
        \llamathree{70}-Instruct & Instruction & 57.3 & 51.2 & 54.3 & 51.8 & 50.6 & 49.4 & 52.4 \\
        \bottomrule
    \end{tabular}}
\end{table}

\subsection{Code Editing and Translation}

\begin{table}
    \caption{Pass@1 and ExcessCode performance on CanItEdit. Pass@1 assesses functional correctness, and ExcessCode assesses conciseness and precision of code edits.
    }
    \label{tab:caniedit}
    \centering
    \small
    \begin{tabular}{ccccc}
        \toprule
        \textbf{Model} & \multicolumn{2}{c}{\textbf{Descriptive}} & \multicolumn{2}{c}{\textbf{Lazy}} \\
        \cmidrule(lr){2-3} \cmidrule(lr){4-5} 
        & \textbf{Pass@1 ($\uparrow$)} & \textbf{ExcessCode ($\downarrow$)} & \textbf{Pass@1 ($\uparrow$)} & \textbf{ExcessCode ($\downarrow$)} \\
        \midrule
        \codegemma{7}-IT & 31.57 & 1.03 & 25.28 & 0.58  \\
        \codellama{7}-Instruct & 33.23 & 0.80 & 24.80 & 0.45 \\
        Granite-8B-Code-Instruct & \textbf{39.72} & \textbf{0.30} & \textbf{32.38} & \textbf{0.08} \\
        \midrule
        \codellama{13}-Instruct & \textbf{41.09} & \textbf{0.20} & \textbf{29.85} & 0.70 \\
        Granite-20B-Code-Instruct & 40.52 & 0.30 & 29.72 & \textbf{0.02} \\
        \midrule
        \codellama{34}-Instruct & 46.28 & \textbf{0.05}  & \textbf{37.90} & 0.43 \\
        Granite-34B-Code-Instruct & \textbf{50.28} & 0.25 & 37.28 & \textbf{0.05} \\
        \bottomrule
    \end{tabular}
\end{table}
\begin{table}[t]

\centering
\caption{Performance of different instruction-tuned models on CodeLingua \citep{pan_lost_2024}. We report Pass@1 for translation across five languages.}

\label{tab:codetlingua}
    \resizebox{1.0\linewidth}{!}{
    \begin{tabular}{c}
\begin{tabular}{crrrrrrrrrrrr}  %
\toprule

{\textbf{Source Language} }& \multicolumn{4}{c}{\textbf{C}}  & \multicolumn{4}{c}{\textbf{C++}} & \multicolumn{4}{c}{\textbf{Go}}    \\ 
\cmidrule(lr){2-5} \cmidrule(lr){6-9}  \cmidrule(lr){10-13} 
{\textbf{Target Language}} & \textbf{C++} & \textbf{Go} & \textbf{Java} & \textbf{Py}& 
\textbf{C} & \textbf{Go} & \textbf{Java} & \textbf{Py}& 
\textbf{C} & \textbf{C++} & \textbf{Java} & \textbf{Py} \\
                       \midrule 
Granite-3B-Code-Instruct & 78.5	&43.5&	64.5&	38.5&	43.5&	25.5&	68&	46.5&	46.5&	62.5&	2.5	&20.0	\\
\midrule
\codegemma{7}-IT & \textbf{79.5}	&\textbf{48.0}&	60.5&	44.0	&29.0&	\textbf{32.0}&	58.0&	39.0&	31.0&	49.5&	55.5&	33.5\\
\codellama{7}-Instruct & 18.0&3.0 &	59.5 &	31.0	&9.0	&13.0&	54.5&	28.0	&5.5	&31.5	&0.0	&5.5 \\
Granite-8B-Code-Instruct & 78.5&	18.5&	\textbf{72.0}	&\textbf{57.0}	&\textbf{54.0}&	24.5&	\textbf{74.0}	&\textbf{56.5}	&\textbf{50.0}	&\textbf{65.5} &	\textbf{58.0} & \textbf{52.5}\\
\midrule
\codellama{13}-Instruct & 2.5	&41.5	&60.0	&2.5&	0.0&	23.0&	59.0&	11.5&	1.5	&2.5	&21.0	&27.5 \\
\octocoder{15} & 1.5 & 39.0 & 1.0 & 1.0 & 0.0 & 29.0 & 4.0 & 0.5 & 0.5 & 6.5 & 0.0 & 37.5 \\
Granite-20B-Code-Instruct & \textbf{81.5}&	\textbf{50.0}	&\textbf{71.5}	&\textbf{39.0}	&\textbf{55.5}	&\textbf{31.5}	&\textbf{70.5}	&\textbf{40.5}	& \textbf{55.0} & \textbf{63.0}	&\textbf{58.0}&	\textbf{55.5}\\
\midrule
\codellama{34}-Instruct  &2.5&	\textbf{49.0}&	60.0&	28.5&	2.0&	\textbf{26.0}&	60.5&	33.0&	0.5&	3.0&	38.5&	18.0 \\
Granite-34B-Code-Instruct & \textbf{83.0}	&15.5	&\textbf{74.5}	&\textbf{54.5}& \textbf{61.0}&	25.0&	\textbf{74.0} &	\textbf{42.5}&	\textbf{58.0}&	\textbf{62.0}&	\textbf{68.0}&	\textbf{66.0}\\
\bottomrule
\end{tabular}
\\
\\
\begin{tabular}{crrrrrrrr}  
\toprule
{\textbf{Source Language}}  & \multicolumn{4}{c}{\textbf{Java}}   & \multicolumn{4}{c}{\textbf{Python}}   \\
\cmidrule(lr){2-5} \cmidrule(lr){6-9}  
{\textbf{Target Language}} & \textbf{C} & \textbf{C++} & \textbf{Go} & \textbf{Py} & 
\textbf{C} & \textbf{C++} & \textbf{Go} & \textbf{Java} \\
\midrule
Granite-3B-Code-IT 	&34.2&	63.2&	20.1&	44.1&	18.8&	45.6&	1.2&	15.0\\
\midrule
\codegemma{7}-IT & 	33.7	&31.3&	10.3&	31.3&	22.9&	33.4&	4.6	&\textbf{44.5}\\
\codellama{7}-Instruct & 	5.3	&17.8	&15.8&	11.6	&2.55	&18.0&	\textbf{14.7}&	28.2\\
Granite-8B-Code-Instruct & \textbf{49.6}&	\textbf{57.3}	&\textbf{17.1}&	\textbf{57.9}	&  \textbf{41.3}	&\textbf{55.2} &	10.5 &	39.7\\
\midrule
\codellama{13}-Instruct &	0.0&	1.25&	7.6&	34.2&	0.0&	0.9&	9.6&	13.1\\
\octocoder{15} & 0.3 & 3.2 & 18.8 & 22.9 & 0.0 & 2.65 & 16.8 & 32.2\\
Granite-20B-Code-Instruct & \textbf{44.8}	& \textbf{52.8}	&\textbf{33.4} & 		\textbf{34.3} &	\textbf{41.8}	&\textbf{48.4}&	\textbf{26.1}&	\textbf{53.4}\\
\midrule
\codellama{34}-Instruct  & 0.0&	1.8&	9.8&	20.6&	0.5&	1.0&	14.1&	10.0 \\
Granite-34B-Code-Instruct & \textbf{55.0}&	\textbf{65.9}	&\textbf{28.5}&	\textbf{56.3}&	\textbf{45.7}&	\textbf{53.7}&	\textbf{34.1}&	\textbf{61.3}	\\

\bottomrule
\end{tabular}
\\
 \end{tabular} \vspace{-6mm}
}
\end{table}

CanItEdit is a recent benchmark designed to evaluate code LLMs on instructional code editing tasks. The benchmark contains 105 hand-crafted Python programs where each problem consists of a code snippet accompanied by
an instruction of two types: descriptive or lazy. The goal is to modify the code according to the instruction; both lazy and descriptive instructions should lead to the same edit. Following~\cite{cassano2024edit}, we compare different instruction-tuned models using their corresponding instruction format, by random sampling with a temperature of 0.2 and a
top-p of 0.95, with 20 completions per problem.

From Table~\ref{tab:caniedit}, we make the following observations on the performance of different models on CanItEdit benchmark. 
It shows that \lm have better pass rates, as well as less presence of unnecessary code changes, compared to CodeGemma and CodeLlama.
This result shows that \lm can better understand users' intentions and make accurate changes to the existing code in practical situations.

CodeLingua \citep{pan_lost_2024} is a dataset designed to test model's capabilities in code translation. It contains two sets of programs: one containing 251 programs in Java and 251 prgrams in Python sampled from Avatar~\citep{ahmad2021avatar}, and one from CodeNet~\cite{puri2021codenet} containing 250 programs for each of five languages: C, C++, Go, Java and Python. For each program a set of unit tests in the form of input and expected outputs is provided. The task consists in translating each program from the source language to five target languages (the ones sampled from CodeNet). Pass@1 is used as the metric to evaluate translation accuracy. For every generation, we used greedy decoding and the suggested prompt format for each instruction tuned model. For base models, or cases where the instruction format was not specified, we used the default prompt from the dataset. Basic post-processing is applied to each generation, to remove generation artifacts such as repetition of the input instruction, source language code, target language name and formatting tokens (\verb|```| , for example).

Table~\ref{tab:codetlingua} shows the results on the CodeLingua benchmark. For the source languages C, C++ and Go the results reported in the table are taken directly from the runs on Codenet, whereas for Java and Python the results are reported as the average of the runs on Avatar and CodeNet. We report the numbers of Octocoder and CodeLlama from the CodeLingua leaderboard~\footnote{\url{https://codetlingua.github.io/leaderboard.html}}.  The Granite Code models performs comparably to CodeGemma. It is worth noticing that the correctness of the translation is not only due to the code generated by the model, but also by the extra metadata and explanation provided as part of the answer. We tested instruction tuned models, as we observed that base models often struggle to understand the request itself to translate code. Instruct models, on the other hand, tend to add additional information besides the translated code as part of the generations. The CodeLLama family seems to suffer especially from this issue, as post-processing the generations to extract only the relevant code constitutes a non-trivial task. The CodeGemma and Granite Models on the other hand, produce a nicely formatted output that can be easily parsed. Interestingly, Go seems to be the hardest target language to translate to, while C is the source language with the highest translation success rate from, for the Granite models. 

\subsection{Code Reasoning, Understanding and Execution}

\begin{table}[ht]
    \caption{Performance on the CRUXEval benchmark. We use temperature 0.2 for pass@1 and temperature 0.8 for pass@5, both using 10 samples, as in~\citep{gu2024cruxeval}.}
    \label{tab:cruxeval}
    \centering
    \small
    \begin{tabular}{c cc cc}
    \toprule
    \multirow{2}*{\textbf{Model}} &
    \multicolumn{2}{c}{\textbf{CRUXEval-I}} &
    \multicolumn{2}{c}{\textbf{CRUXEval-O}}
    \\ \cmidrule(lr){2-3} \cmidrule(lr){4-5}
    & \textbf{Pass\@1} & \textbf{Pass\@5} 
    & \textbf{Pass\@1} & \textbf{Pass\@5}  \\
    \midrule
    \starcoderbase{3} & 27.5 & 44.9 & 27.0 & 41.8\\
    \stablecode{3} & \textbf{33.5} & \textbf{54.2} & \textbf{32.1} & \textbf{44.0} \\
    \starcodertwo{3} & 32.1  & 50.3 & 34.0 & 48.4 \\
    \codegemma{2} & 29.6 & 45.5 & 32.1 & 45.5 \\
    Granite-3B-Code-Base & 30.6 & 50.9 & 31.4 & 35.2 \\

    \midrule
    \starcoderbase{7}  & 29.8 & 47.5 & 32.1 & 44.2 \\
    \codellama{7} & 36.2 & 53.7 & 34.0 & 48.5 \\
    \starcodertwo{7} & 34.2 & 53.8 & 35.8 & 49.7 \\
    \codegemma{7} & \textbf{42.6} & \textbf{60.9} & \textbf{43.9} & \textbf{56.7} \\
    Granite-8B-Code-Base & 36.0 & 55.8 & 36.1 & 50.3 \\
    \midrule
    \starcoderbase{15} & 31.0 & 49.2 & 34.4 & 47.4 \\
    \codellama{13}   & 42.2 & 61.8 & 39.9 & 55.1 \\
    \starcodertwo{15}  & \textbf{47.4} & \textbf{68.3} & \textbf{46.7} & \textbf{59.2} \\
    Granite-20B-Code-Base & 39.1 & 59.0 & 37.5 & 51.7 \\
    \midrule
    \codellama{34} & \textbf{47.8} & \textbf{65.6} & 42.5 & \textbf{56.5} \\
    Granite-34B-Code-Base & 43.3 & 61.3 & \textbf{44.8} & 55.1 \\
    \bottomrule
    \end{tabular} 
\end{table}

CRUXEval~\citep{gu2024cruxeval} is a benchmark of 800 Python functions and input-output pairs, consisting of two tasks: CRUXEval-I (input prediction) and CRUXEval-O (output prediction). We use temperature 0.2 to report pass@1 and temperature 0.8 to report pass@5, both using 10 samples, as in~\cite{lozhkov2024starcoder,gu2024cruxeval}. Table~\ref{tab:cruxeval} shows that Granite Code models perform competitively with other models. Granite-3B-Code-Base outperforms CodeGemma-2B on CRUXEval-I but lags behind on CRUXEval-O. Interestingly, there is not a single model which performs consistently best at 3B parameters. However, at 7B-8B parameters, CodeGemma-7B outperforms all the models on both tasks. For the large models, Granite-34B-Code-Base lags behind CodeLlama-34B on CRUXEval-I but outperforms on CRUXEval-O. Performance on both CRUXEval-I and CRUXEval-O increases as we scale the size of the Granite Code models from 3B to 34B parameters, demonstrating the advantage of larger models for code reasoning and execution tasks.  

\subsection{Math Reasoning}

\begin{table}[ht]
    \caption{Performance on 4 chain-of-thought math tasks and 2 tool-aided math tasks.}
    \label{tab:reasoning}
    \centering
    \small
    \begin{tabular}{ccccccc}
    \toprule
        \textbf{Model} & \textbf{MATH} & \textbf{GSM8K} & \textbf{SAT} & \textbf{OCW} & \textbf{MATH+Py} & \textbf{GSM8K+Py} \\ 
        \midrule
        \starcoderbase{7} & 2.4 & 3.8 & 18.7 & 2.2 & 18.2 & 15.6 \\
        \codellama{7}  & 4.1 & 11.9 & 12.5 & 2.9 & 20.8 & 26.8 \\ 
        \starcodertwo{7}  & 10.4 & 27.2 & 37.5 & 4.8 & 28.7 & 39.4 \\ 
        \codegemma{7} & 21.8 & 49.0 & 53.1 & 6.9 & 31.1 & 60.9 \\
        Granite-8B-Code-Base & 21.4 & \textbf{61.9} & 62.5 & 8.8 & \textbf{35.4} & \textbf{63.1} \\
        \midrule
        \gemma{7} & \textbf{24.1} & 53.3 & \textbf{75.0} & 7.3 & 27.4 & 52.9 \\
        \mistral{7}-v0.2 & 12.8 & 37.2 & 53.1 & 5.8 & 25.7 & 45.6 \\
        \llamathree{8} & 15.6 & 49.8 & 34.4 & \textbf{9.9} & \hspace{1.5mm}0.0$^\star$& \hspace{1.5mm} 2.4$^\star$ \\
        \llemma{7} & 17.3 & 33.7 & 59.4 & 7.0 & 25.6 & 40.8 \\
        \bottomrule \\
        \multicolumn{7}{l}{$^\star$ We noticed that Llama-3-8B-Base tends to generate invalid programs given the same prompts}  \\
        \multicolumn{7}{l}{\hspace{1.8mm}   as the other model, resulting in very low scores on MATH+Py and GSM8K+Py tasks. Similar} \\
        \multicolumn{7}{l}{\hspace{1.8mm}  issues have been observed in our Python code generation experiment.}
    \end{tabular}
\end{table}

\begin{figure} [h]
\begin{center}
\includegraphics[width=0.9\linewidth]{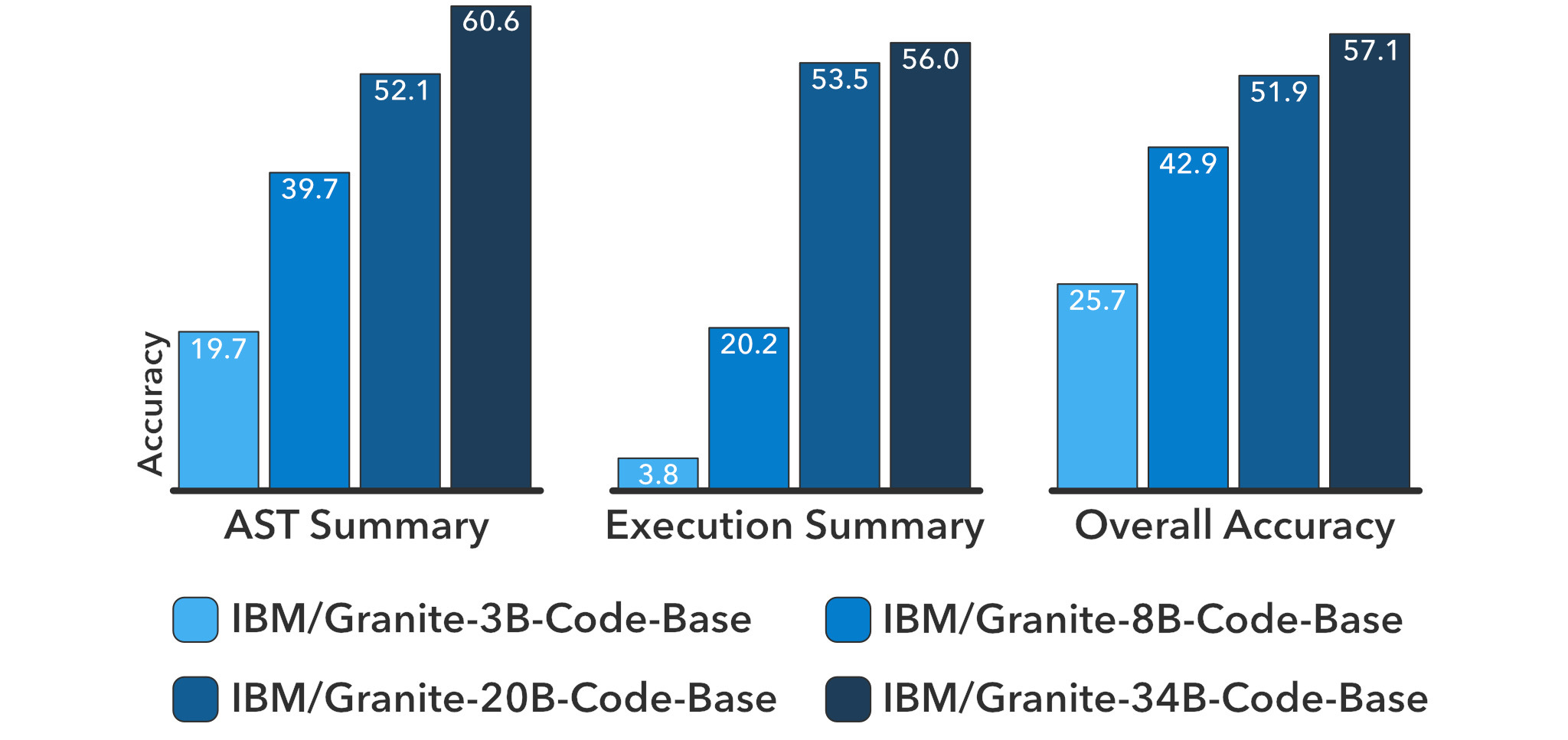}
\end{center} 
\caption{Performance of Granite Code models on Berkeley Function-Calling Leaderboard. Overall accuracy keeps increasing with increase in model sizes, indicating the advantage of large models for function calling abilities. Best viewed in color.}
\label{fig:function-granite}
\end{figure}

We use the following four widely used benchmarks to assess the mathematical reasoning capabilities of Granite-8B-Code-Base and various 7B-8B baseline models:
\begin{itemize}
    \item \textbf{MATH}~\citep{hendrycksmath2021}: a dataset from high-school math competitions; we use the 4-shot experiment setting from \cite{eval-harness};
    \item \textbf{GSM8K}~\citep{cobbe2021training}: a dataset of middle-school level math word problems; we use the 5-shot experiment setting from \cite{eval-harness};
    \item \textbf{SAT}~\citep{azerbayev2023llemma}: a dataset consisting of the 32 math questions with no figures from the May 2023 College Board SAT examination; we use the same experiment setting from \cite{azerbayev2023llemma}
    \item \textbf{OCW}~\citep{lewkowycz2022solving}:  a collection of undergraduate-level STEM problems harvested from MIT’s OpenCourseWare; we use the 4-shot experiment setting from \cite{azerbayev2023llemma}.
\end{itemize}

Following \cite{azerbayev2023llemma}, we also evaluate models on two tasks that involve solving problems with access to computational tools:
\begin{itemize}
    \item \textbf{MATH+Py} solving MATH task by writing a Python program that uses built-in numeric operations, the \texttt{math} module, and \texttt{SymPy}; we use the 5-shot prompt and experiment setting from \cite{azerbayev2023llemma};
    \item \textbf{GSM8K+Py} solving GSM8K task by writing a Python program that executes to generate an integer answer; we use the 8-shot prompt and experiment setting from \cite{azerbayev2023llemma}.
\end{itemize}

Table~\ref{tab:reasoning} summarizes the results. 
Despite not being specifically tuned for mathematical reasoning, Granite-8B-Code-Base shows impressive reasoning ability, outperforming most existing 7B to 8B models.
While other models may be particularly strong on a few tasks, our model consistently achieves top-1 or top-2 performance on all tasks.

\subsection{Calling Functions and Tools}
\label{sec:function}

We adopt Berkeley Function-Calling Leaderboard (BFCL)~\citep{berkeley-function-calling-leaderboard}, to evaluate LLM's ability to call functions and tools.
BFCL is a function-calling dataset with 1700 functions across 4 categories: simple, multiple, parallel, and parallel multiple function calls - each differing in the number of potential functions the model has access to and the number of output functions the model has to generate.
We use two popular methods to evaluate the accuracy of the model-generated answers: AST evaluation based on Abstract Syntax Tree (AST) based metric for fuzzy evaluation of output, and Executable evaluation to match the outputs of model-generated and ground-truth functions.

\begin{figure} [t]
\begin{center}
\includegraphics[width=0.9\linewidth]{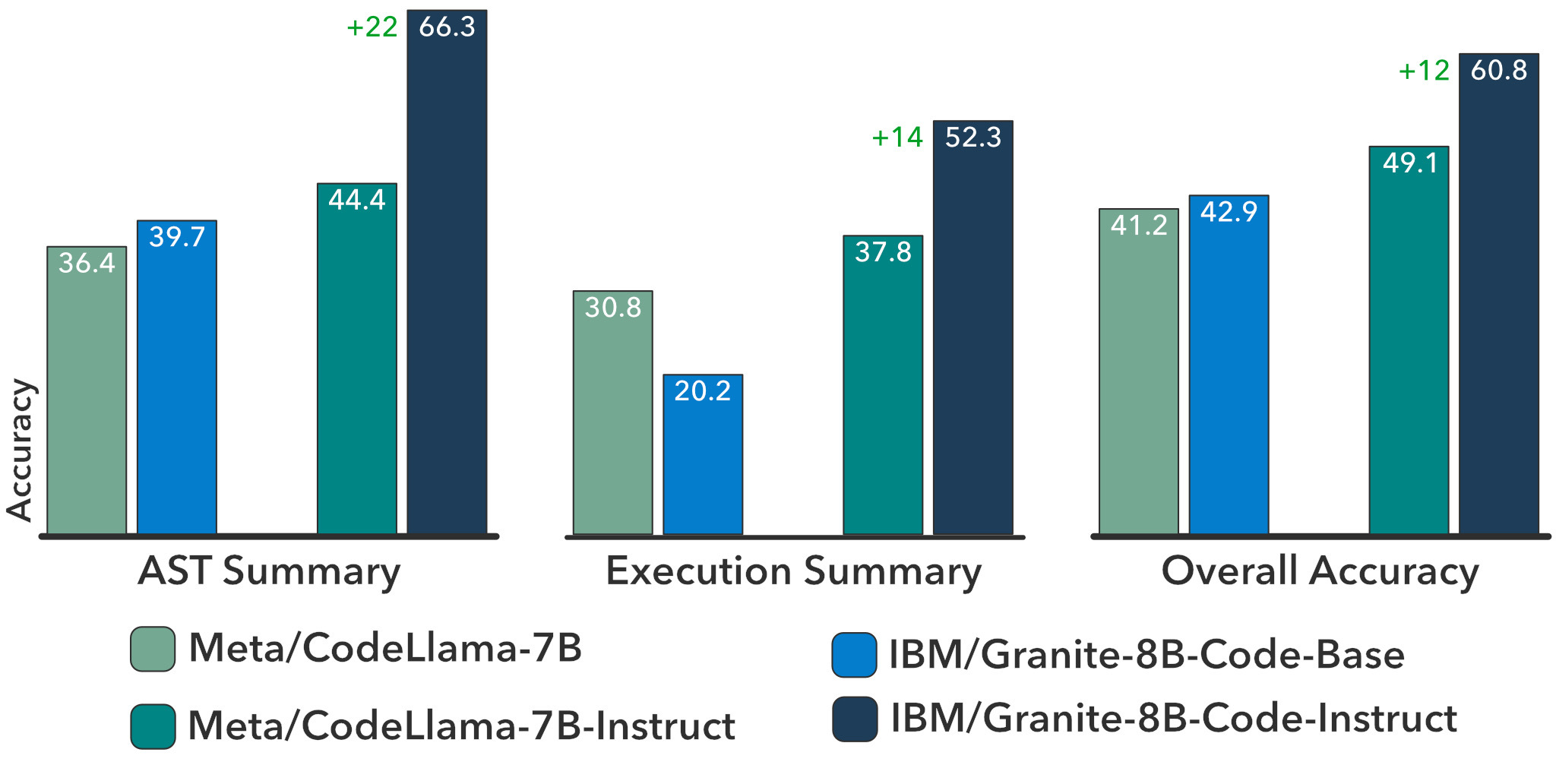}
\end{center} 
\caption{Granite-8B-Code vs CodeLlama-7B on Berkley Function-Calling Leaderboard. Granite-8B-Code (Base/Instruct) consistently outperforms CodeLlama-7B (Base/Instruct) on all three metrics. Best viewed in color.}
\label{fig:function-granite-llama} \vspace{-2mm}
\end{figure}

Figure~\ref{fig:function-granite} shows the results of different Granite Code models on BFCL benchmark. As can be seen from the figure, overall accuracy improves from 25.65\% to 57.12\% for Granite-3B-Code-Base to Granite-34B-Code-Base, showing the effectiveness of model scaling in function (tool) calling capabilities. We also compare Granite-8B-Code with CodeLlama-7B in Figure~\ref{fig:function-granite-llama} and find that Granite-8B-Code-Instruct beats CodeLlama-7B-Instruct by 22\%, 14\% and 12\% on AST Summary, Execution Summary and Overall accuracy respectively. Additionally, Figure~\ref{fig:function-granite-llama} shows that instruction tuning consistently improves performance of both base models, with more noticeable improvements in Granite Code models. E.g., +17.88\% in overall accuracy from Granite-8B-Code-Base to Granite-8B-Code-Instruct, indicating the effectiveness of our well-curated data mixture in finetuning base models. 

\vspace{-2mm}
\subsection{Model Robustness}

\begin{table}[t]
    \caption{RP@1 performance on the Recode benchmark. Following~\citep{recode_wang2022}, we use the perturbed version of the HumanEval benchmark with greedy sampling for all the models to eliminate randomness effect and enable fair comparison.} 
    \label{tab:recode_benchmark}
    \centering
    \small
    \begin{tabular}{ccccc}
    \toprule
    \textbf{Model}             & \textbf{Docstring}             & \textbf{Function} & \textbf{Syntax}             & \textbf{Format}         \\
    \midrule
    \starcoderbase{3} & 12.3 & 11.4 & 17.2 & 24.2 \\
    \stablecode{3} & 22.8 & 25.8 & 37.1 & 46.4 \\
    \starcodertwo{3} & \textbf{28.6} & 29.7 & \textbf{49.6} & \textbf{57.6}  \\
    \codegemma{2} & 12.3 & 11.4 & 17.2 & 24.2 \\
    Granite-3B-Code-Base  & 28.2 & \textbf{30.0} & 45.8 & 56.3 \\
    \midrule
    \starcoderbase{7} & 23.7 & 25.3 & 38.2 & 47.1 \\
    \codellama{7} & 24.7 & 27.6 & 43.0 & 53.1 \\
    \starcodertwo{7} & 27.6 & 30.4 & 45.8 & 57.5 \\
    \codegemma{7} & \textbf{32.3} & \textbf{37.8} & \textbf{55.3} & \textbf{64.3} \\
    Granite-8B-Code-Base & 25.5 & 30.9 & 49.9 & 60.5 \\
    \midrule
    \starcoderbase{15}  & 26.6 & 30.7 & 44.3 & 52.2 \\
    \codellama{13} & 25.8 & 29.7 & 50.6 & 60.3 \\
    \starcodertwo{15} & \textbf{36.9} & \textbf{43.9} & \textbf{60.4} & \textbf{70.2} \\
    Granite-20B-Code-Base & 35.2 & 43.0 & 55.1 & 63.5  \\
    \midrule
    \codellama{34} & 33.1 & 38.0 & 54.7 & 64.4 \\
    Granite-34B-Code-Base & \textbf{36.3} & \textbf{44.4} & \textbf{59.2} & \textbf{68.2} \\
    \bottomrule
    \end{tabular}
\end{table}

While the performance on canonical code generative tasks is essential, we argue that the evaluation of practical robustness is also necessary to characterize different models systematically. We therefore
consider benchmarking the robustness of code synthesis, one of the most representative downstream
tasks of source code. ReCode~\citep{recode_wang2022} provides 30 different general perturbations on docstrings, function names, and codes to evaluate the  robustness of code-generation models. We use the perturbed version of the HumanEval benchmark using greedy generation with 5 seeds, as recommended in~\citep{recode_wang2022}. 

Table~\ref{tab:recode_benchmark} shows the worst-case RP@1 of different models for each perturbation category. While Granite-3B-Code-Base consistently outperforms CodeGemma-2B, Granite-8B-Code-Base lags behind CodeGemma-7B on all categories. Granite Code models obtains much better performance compared to CodeLlama models, showing its generalization in a robust way at every sizes. Our largest model, Granite-34B-Code-Base consistently outperforms CodeLlama-34B on all four categories. This indicates that Granite-34B-Code-Base has more capacity to deal with unseen instances and perturbations. In general, we also observe higher RP@1 for larger models within the Granite Code family (e.g., improved from 40.1\% to 52.0\% for Granite-3B-Code-Base to Granite-34B-Code-Base on average across all perturbations), showing that larger model helps improve worst-case robustness.  

\section{Conclusion}
\label{sec:conclusion}

We presented a family of decoder-only \lm ranging in size from 3 to 34 billion parameters that are highly versatile in their ability to accomplish a wide range of tasks from code generation to fixing bugs, explaining and documenting code, maintaining repositories, and more. These models have proven to be suitable for applications ranging from complex application modernization tasks~\citep{wca} to on-device memory-constrained use cases. Extensive evaluation demonstrates that \lm consistently reach state-of-the-art performance among open-source code LLMs, matching or exceeding the performance of recently released CodeGemma, StarCoder2, and Llama3 models on average performance across various code-related tasks of code generation, explanation, and bug fixing in a variety of popular programming languages. Our experience and results demonstrate that \lm have a proven ability to better handle different tasks in enterprise software development workflows.
We release all our \lm under an Apache 2.0 license for both research and commercial use.
We plan to continuously release updates to these models to improve their performance, e.g. leveraging the CodeNet instruction dataset~\citep{puri2021codenet}, and in the near future we plan to release long-context as well as Python- and Java-specialized model variants.
\section*{Acknowledgments}

We would like to acknowledge the efforts of numerous teams at IBM Research AI and Hybrid Cloud Platform, IBM AI Infrastructure team, IBM WatsonX Code Assistant and platform team. Special thanks to IBM Research leaders - Dario Gil, Sriram Raghavan, Mukesh Khare, Danny Barnett, Talia Gershon, Priya Nagpurkar, Nicholas Fuller for their support.

Thanks and acknowledgement to Trent Gray-Donald, Keri Olson, Alvin Tan, Hillery Hunter, Dakshi Agrawal, Xuan Liu, Mudhakar Srivatsa, Raghu Kiran Ganti, Carlos Costa, Darrell Reimer, Maja Vukovic, Dinesh Garg, Akash Srivastava, Abhishek Bhandwaldar, Aldo Pareja, Shiv Sudalairaj, Atin Sood, Sandeep Gopisetty, Nick Hill, Ray Rose, Tulio Coppola, Állysson Oliveira, Aadarsh Sahoo, Apoorve Mohan, Yuan Chi Chang, Jitendra Singh, Yuya Ong, Eric Butler, David Brotherton, Rakesh Mohan, David Kung, Dinesh Khandelwal, Naigang Wang, Nelson Mimura Gonzalez, Olivier Tardieu, Tuan Hoang Trong, Luis Angel Bathen, Kevin O'Connor, Christopher Laibinis, Tatsuhiro Chiba, Sunyanan Choochotkaew, Robert Walkup, Antoni Viros i Martin, Adnan Hoque, Davis Wertheimer and Marquita Ellis.

\newpage
\bibliography{0_main}
\bibliographystyle{colm2024_conference}

\newpage
\appendix
\section{Programming Languages}
\label{sec:lang}

ABAP, Ada, Agda, Alloy, ANTLR, AppleScript, Arduino, ASP, Assembly, Augeas, Awk, Batchfile, Bison, Bluespec, C, C-sharp, C++, Clojure, CMake, COBOL, CoffeeScript, Common-Lisp, CSS, Cucumber, Cuda, Cython, Dart, Dockerfile, Eagle, Elixir, Elm, Emacs-Lisp, Erlang, F-sharp, FORTRAN, GLSL, GO, Gradle, GraphQL, Groovy, Haskell, Haxe, HCL, HTML, Idris, Isabelle, Java, Java-Server-Pages, JavaScript, JSON, JSON5, JSONiq, JSONLD, JSX, Julia, Jupyter, Kotlin, Lean, Literate-Agda, Literate-CoffeeScript, Literate-Haskell, Lua, Makefile, Maple, Markdown, Mathematica, Matlab, Objective-C++, OCaml, OpenCL, Pascal, Perl, PHP, PowerShell, Prolog, Protocol-Buffer, Python, Python-traceback, R, Racket, RDoc, Restructuredtext, RHTML, RMarkdown, Ruby, Rust, SAS, Scala, Scheme, Shell, Smalltalk, Solidity, SPARQL, SQL, Stan, Standard-ML, Stata, Swift, SystemVerilog, Tcl, Tcsh, Tex, Thrift, Twig, TypeScript, Verilog, VHDL, Visual-Basic, Vue, Web-Ontology-Language, WebAssembly, XML, XSLT, Yacc, YAML, Zig

\end{document}